%% file: diversity-icml.tex
\documentclass{article}

\usepackage{microtype}
\usepackage{graphicx}
\usepackage{subfigure}
\usepackage{booktabs} 
\usepackage[table]{xcolor}
\usepackage[toc,page,header]{appendix}
\usepackage{pdfpages}
\usepackage{dsfont}

\input{math_commands}

\usepackage{amssymb}
\usepackage{bbm}

\newtheorem{theorem}{Theorem}
\newtheorem{definition}[theorem]{Definition}
\newtheorem{proposition}[theorem]{Proposition}

\makeatletter
\newenvironment{smalleralign}[1][\small]
 {\par\nopagebreak\leavevmode\vspace*{-\baselineskip}%
  \skip0=\abovedisplayskip
  #1%
  \def\maketag@@@##1{\hbox{\m@th\normalfont\normalsize##1}}%
  \abovedisplayskip=\skip0
  \align}
 {\endalign\ignorespacesafterend}
\makeatother

\usepackage{hyperref}

\usepackage[accepted]{icml2021}

\icmltitlerunning{Modelling Behavioural  Diversity for  Learning in Open-Ended Games}
\newcolumntype{C}[1]{>{\centering\let\newline\\\arraybackslash\hspace{0pt}}m{#1}}

\begin{document}

\setlength{\abovedisplayskip}{3pt}
\setlength{\belowdisplayskip}{2pt}

\twocolumn[
\icmltitle{Modelling Behavioural  Diversity for  Learning in Open-Ended Games}



\icmlsetsymbol{equal}{*}

\begin{icmlauthorlist}
\icmlauthor{Nicolas Perez Nieves}{equal,huawei,ic}
\icmlauthor{Yaodong Yang}{equal,huawei,ucl}
\icmlauthor{Oliver Slumbers}{equal,ucl}
\icmlauthor{David Henry Mguni}{huawei}
\icmlauthor{Ying Wen}{ucl}
\icmlauthor{Jun Wang}{huawei,ucl}
\end{icmlauthorlist}

\icmlaffiliation{huawei}{Huawei U.K.}
\icmlaffiliation{ucl}{University College London}
\icmlaffiliation{ic}{Imperial College London, work done during  internship at Huawei U.K.}

\icmlcorrespondingauthor{}{yaodong.yang@outlook.com}

\icmlkeywords{Machine Learning, ICML}

\vskip 0.3in
]



\printAffiliationsAndNotice{\icmlEqualContribution} 

\begin{abstract}

Promoting behavioural diversity is critical for solving games with non-transitive dynamics where strategic cycles exist, and there is no consistent winner (e.g., Rock-Paper-Scissors). Yet, there is a lack of rigorous treatment for defining diversity and  constructing diversity-aware  learning dynamics. In this work, we offer a  geometric interpretation of  behavioural diversity in games and introduce a novel diversity metric based on \emph{determinantal point processes} (DPP). By incorporating the diversity metric into  best-response dynamics, we develop \emph{diverse fictitious play} and \emph{diverse policy-space response oracle} for solving normal-form games and open-ended games. We prove  the uniqueness of the diverse best response and the convergence of our algorithms on two-player games. Importantly, we show that maximising the DPP-based diversity metric   guarantees to enlarge the \emph{gamescape} -- convex polytopes spanned by agents' mixtures of strategies. To validate our diversity-aware solvers, we test on tens of games that show  strong non-transitivity. Results suggest that  our methods  achieve at least the same, and in most games, lower exploitability than PSRO   solvers by finding effective and diverse strategies. 
\end{abstract}

\vspace{-15pt}
\section{Introduction}
\vspace{-3pt}
Nature exhibits a remarkable tendency towards \textit{diversity}  \citep{holland1992adaptation}. 
Over the past billions of years, natural evolution has discovered a vast assortment of unique species. Each of them is capable of orchestrating, in different ways,  the complex biological processes that are necessary to sustain life. 
Equally, in computer science, machine intelligence can be considered as the ability to adapt to a diverse set of complex environments \citep{hernandez2017measure}.  
This suggests that the intelligence of AI evolves with environments of increasing diversity.  In fact, 
recent successes in developing AIs that achieve super-human performance on  sophisticated  battle games \citep{vinyals2019grandmaster,ye2020towards}  have  provided factual justifications for promoting behavioural diversity in training  intelligent  agents.

In game theory, the necessity of pursuing behavioural  diversity is also deeply rooted in the non-transitive structure of games \citep{balduzzi2019open}. 
In general, an arbitrary game, of either the normal-form type \citep{candogan2011flows} or the differential type \citep{balduzzi2018mechanics}, can always be decomposed into a sum of two components: a \emph{transitive part} and a \emph{non-transitive part}. 
The transitive part of a game represents the structure in which the rule of winning is transitive (i.e., if strategy A beats B, B beats C, then A beats C), and the non-transitive part refers to the  structure in which the set of strategies follows a cyclic rule (e.g., the endless cycles among Rock, Paper and Scissors). 
Diversity matters especially for the non-transitive part simply because there is no consistent winner in such part of a game: if a player only plays Rock, he can be \emph{exploited} by Paper, but not so if he has a diverse strategy set of    Rock and Scissor.

In fact, many real-world games demonstrate strong non-transitivity  \cite{czarnecki2020real}; 
therefore, it is critical to design objectives  in the learning framework that can lead to behavioural diversity. 
In multi-agent reinforcement learning (MARL) \cite{yang2020overview},  
promoting  diversity not only prevents AI agents from checking the same policies repeatedly, but more importantly, helps them  discover niche skills, avoid being exploited and maintain robust performance when encountering unfamiliar types of  opponents.    
In the examples of   StarCraft \cite{vinyals2019grandmaster},  Soccer \cite{kurach2020google} and autonomous driving \cite{zhou2020smarts}, learning a diverse set of  strategies has been reported as an imperative step in strengthening AI's performance.


Despite the importance of diversity \cite{yang2021diverse}, there is very little work that  offers a  rigorous treatment in even defining diversity.   
The majority of work so far has followed a heuristic approach. For example, 
the idea of \emph{co-evolution} \citep{durham1991coevolution,paredis1995coevolutionary} has drawn forth a series of effective methods,  
such as open-ended evolution \citep{standish2003open, banzhaf2016defining,lehman2008exploiting}, population based training methods \citep{jaderberg2019human,liu2018emergent}, and auto-curricula \citep{leibo2019autocurricula, baker2019emergent}. 
Despite many empirical successes,    the lack of rigorous treatment for   behavioural diversity still  hinders one from developing a principled approach.  


In this work, we introduce a rigorous way of modelling behavioural   diversity for learning in games. Our approach offers a new geometric interpretation, which is built upon \emph{determinantal point processes} (DPP) that have origins in modelling repulsive quantum particles  \citep{macchi1977fermion} in physics.  
A DPP is a special type of point process, which measures the probability of selecting a random subset from a ground set where only diverse subsets are desired. 
We adapt DPPs to games by formulating the \emph{expected cardinality} of a DPP as the diversity metric. 
The proposed diversity metric is a general tool for game solvers; we incorporate our diversity metric into the best-response dynamics, and develop  diversity-aware extensions of \emph{fictitious play} (FP) \citep{brown1951iterative} and \emph{policy-space response oracles} (PSRO) \citep{lanctot2017unified}. 
Theoretically, 
we show that maximising the DPP-based diversity metric guarantees an expansion of the gamescape spanned by agents' mixtures of policies.
Meanwhile, we prove the convergence of our diversity-aware learning methods to the respective solution concept of Nash equilibrium  and $\alpha$-Rank \citep{omidshafiei2019alpha} in two-player games. 
Empirically, we evaluate our methods on tens of games that show strong non-transitivity, covering both normal-form games and open-ended games. 
Results confirm the superior performance of  our methods, in terms of lower exploitability,  against the state-of-the-art game solvers. 

\vspace{-3pt}
\section{Related Work}
\vspace{-3pt}
Diversity has been extensively studied in  evolutionary computation (EC) \citep{fogel2006evolutionary} where the central focus is mimicking the   natural evolution process. 
 One classic idea in EC is \emph{novelty search}  \citep{lehman2011abandoning}, which  searches   for models that lead to different outcomes.  
 Quality-diversity (QD) \citep{pugh2016quality} hybridises novelty search with a fitness objective; two resulting methods are \emph{Novelty Search with Local Competition} \citep{lehman2011evolving} and  \emph{MAP-Elites} \citep{mouret2015illuminating}. 
For solving games, QD methods were applied to ensure policy diversification among learning agents  \citep{gangwani2020harnessing, banzhaf2016defining}. 
Despite  remarkable successes   \citep{jaderberg2019human,cully2015robots},  quantifying diversity in   EC is often task-dependent and hand-crafted; as a result, building a theoretical understanding of how diversity is generated  during learning  is non-trivial  \citep{brown2005diversity}.

Searching for behavioural diversity is  also a common topic in reinforcement learning (RL). Specifically, it is studied under the names of skill discovery \citep{eysenbach2018diversity,hausman2018learning}, intrinsic exploration  \citep{gregor2017variational,bellemare2016unifying,barto2013intrinsic}, or  maximum-entropy learning \citep{haarnoja2017reinforcement,haarnoja2018soft,levine2018reinforcement}. 
These solutions  can still be regarded as QD methods, in the sense that the quality refers to the cumulative reward, and dependent on the context, diversity could refer to policies that visit new states \citep{eysenbach2018diversity} or have a large entropy  \citep{levine2018reinforcement}. 
Two related works in RL, yet with a different scope, are Q-DPP \citep{yang2020multi}, which adopts DPP  to factorise agents' joint Q-functions in MARL, and DvD \citep{parker2020effective}, which studies diversity based on the ensembles of policy embeddings. 

For two-player zero-sum games, 
smooth FP  \citep{fudenberg1995consistency} is a solver that accounts for diversity through adopting a policy entropy term in the original FP \citep{brown1951iterative}. 
When the game size is large, 
\emph{Double Oracle} (DO) \citep{mcmahan2003planning} provides an iterative method  where  agents progressively expand their policy pool by, at each iteration, adding one  best response  versus the opponent's Nash strategy.  
Online DO \citep{dinh2021online} considers a no-regret best response. 
PSRO generalises FP and DO via   
 adopting a RL subroutine to approximate the best response \citep{lanctot2017unified}.   
 Pipeline-PSRO \citep{mcaleer2020pipeline} trains multiple best responses in parallel and efficiently solves games of size $10^{50}$. 
 PSRO$_{rN}$  \cite{balduzzi2019open} is a specific variation of PSRO that accounts for  diversity; however, it suffers from  poor  performance in a selection of tasks \citep{muller2019generalized}. 
Since computing NE is PPAD-Hard \citep{daskalakis2009complexity}, 
another important extension of PSRO is  \emph{$\alpha$-PSRO}  \cite{muller2019generalized}, which replaces NE with \emph{$\alpha$-Rank} \citep{omidshafiei2019alpha,yang2020alphaalpha}, a  solution concept that has  polynomial-time solutions on general-sum games. 
 Yet, how to promote  diversity in the context of $\alpha$-PSRO is still unknown.  
 In this work,  we develop diversity-aware extensions of FP, PSRO and $\alpha$-PSRO, and show on tens of games that our  diverse  solvers achieve  significantly  lower exploitability than the non-diverse baselines.  

\vspace{-3pt}
\section{Notations \& Preliminary}
\vspace{-3pt}
We consider normal-form games (NFGs), denoted by $\langle \mathcal{N}, \mathbb{S} , \mG \rangle$, where each player $i \in \mathcal{N}$ has a finite set of pure strategies $\mathbb{S}^i$. 
Let $\mathbb{S} = \prod_{i\in \mathcal{N}} \mathbb{S}^i$ denote the space of joint pure-strategy profiles, and $\mathbb{S}^{-i}$ denote the set of joint strategy profiles except the $i$-th player. 
A mixed strategy of player $i$ is written by $\vpi^i \in \Delta_{\mathbb{S}^i}$ where $\Delta$ is a probability simplex. A joint mixed-strategy profile is $\boldsymbol{\pi} \in \Delta_{\mathbb{S}}$, and $\boldsymbol{\pi}(S) = \prod_{i \in \mathcal{N}} \vpi^i(S^i)$ represents the probability of joint strategy profile $S$. 
For each $S \in \mathbb{S}$, let $\mG(S) = \big(\mG^1(S), ..., \mG^N(S)\big) \in \mathbb{R}^N$ denote the vector of payoff values for each player. The expected payoff of player $i$ under a joint mixed-strategy profile $\boldsymbol{\pi}$ is thus written as $\mG^i(\boldsymbol{\pi}) = \sum_{S \in \mathbb{S}} \boldsymbol{\pi}(S) \mG^i(S)$, also  as $\mG^i(\boldsymbol{\pi}^i, \boldsymbol{\pi}^{-i})$. 

\vspace{-5pt}
\subsection{Solution Concepts of Games}
\vspace{-5pt}
Nash equilibrium (NE)  exists in all finite games \citep{nash1950equilibrium}; it  is a joint mixed-strategy profile $\vpi$ in which each player $i \in \mathcal{N}$ plays the  \emph{best response} to other players s.t. $\vpi^i \in \mathbf{BR}^i(\vpi^{-i}) := \arg \max_{\vpi \in \Delta_{S^i}}\big[\mG^i(\vpi, \vpi^{-i})\big] $.  For $\epsilon > 0$, an $\epsilon$-best response to the  $\vpi^{-i}$ is 
$\mathbf{BR}_\epsilon^i(\vpi^{-i}) :=\big\{\vpi^i: \mG^i\big(\vpi^i, \vpi^{-i} \big) \ge \mG^i \big(\mathbf{BR}^i(\vpi^{-i}) , \vpi^{-i}\big)  - \epsilon \big\}$, and an $\epsilon$-NE is a joint profile $\vpi \text{ s.t. } \vpi^i \in \mathbf{BR}_\epsilon^i(\vpi),  \forall i \in \mathcal{N}$. 
The \emph{exploitability}  \cite{davis2014using} measures the distance of a joint strategy profile $\vpi$ to a NE, written as 
{\begin{smalleralign}[\small]
\operatorname{Exploit.}\big(\vpi\big) = \sum_{i \in \mathcal{N}}	\Big[\mG^i\big(\mathbf{BR}^i(\vpi^{-i}), \vpi^{-i}\big) - \mG^i\big(\vpi\big)  \Big]. 
\label{eq:nashconv}
\end{smalleralign}}
\vspace{-8pt}

When the exploitability reaches zero, all players reach their best responses,  and thus $\vpi$ is a NE. 

Computing NE in multi-player general-sum games is PPAD-Hard \citep{daskalakis2009complexity}. No polynomial-time solution is available even in two-player cases \citep{chen2009settling}. Additionally,  NE may not be unique.
$\alpha$-Rank \citep{omidshafiei2019alpha} is an alternative  solution concept, 
which  is built on the \emph{response graph} of a game. 
Specifically, $\alpha$-Rank defines the so-called  \emph{sink strongly-connected components} (SSCC) nodes on the response graph that have only incoming edges but no outgoing edges.
The SSCC of $\alpha$-Rank  serves as a promising  replacement for NE; the key associated benefits are its uniqueness, and its polynomial-time solvability in $\mathcal{N}$-player general-sum games.
A more detailed description of $\alpha$-Rank can be found in Appendix A. 

\vspace{-5pt}
\subsection{Open-Ended Meta-Games}
\vspace{-5pt}
The framework of NFGs is often limited in describing real-world games. In solving  games like StarCraft or GO, it is inefficient to list all atomic actions; instead, we are more interested in games at the policy level where a policy can be a ``higher-level" strategy (e.g., a RL model powered by a DNN), and the resulting game is a \emph{meta-game}, denoted by $\langle   \mathcal{N}, \mathbb{S} , \tM\rangle$. 
A meta-game payoff table $\tM$ is  constructed by simulating games that cover different policy combinations. 
With slight abuse of notation\footnote{NFGs and  meta-games are different by the payoff $\mG$ vs.  $\tM$. }, in meta-games, we  respectively use $\mathbb{S}^i$ to denote the  policy set (e.g.,  a population of deep RL models), and use $\vpi^i \in \Delta_{\mathbb{S}^i}$ to denote the meta-policy (e.g., player $i$ plays [RL-Model 1, RL-Model 2] with probability [0.3, 0.7]), and thus  $\vpi=(\vpi^1, ..., \vpi^N)$ is a joint meta-policy profile.  Meta-games are often \emph{\textbf{open-ended}} because there could exist an infinite number of  policies to  play a game. The openness also refers to the fact that 
new strategies will be continuously  discovered and added to agents' policy sets during training; the dimension of $\tM$ will grow.

In the meta-game analysis (\emph{a.k.a.} empirical game-theoretic analysis) \citep{wellman2006methods,tuyls2018generalised}, traditional solution concepts (e.g., NE or $\alpha$-Rank) can still be computed based on $\tM$, even in a more scalable manner, this is because the number of ``higher-level" strategies in the meta-game is usually far smaller than the number of atomic actions of the underlying game.  
For example, in tackling StarCraft \cite{vinyals2019alphastar}, hundreds of deep RL models were trained, which is a trivial amount compared to the number of atomic actions: $10^{26}$ at every time-step.   

Many real-world games (e.g., Poker, GO and StarCraft)  can be described through an open-ended  zero-sum meta-game. 
Given a game engine $\phi: \mathbb{S}^1 \times \mathbb{S}^2 \rightarrow \mathbb{R}$ where $\phi(S^1, S^2)> 0$ if $S^1 \in \mathbb{S}^1$ beats $S^2 \in \mathbb{S}^2$, and $\phi < 0, \phi=0$ refers to losses and ties, 
 the meta-game payoff is  
\begin{smalleralign}[\small]
\tM=\big\{\phi(S^1, S^2): (S^1, S^2) \in \mathbb{S}^1 \times \mathbb{S}^2 \big\}.  
\end{smalleralign}
\vspace{-15pt}

A game is \emph{\textbf{symmetric}} if $\mathbb{S}^1\!=\!\mathbb{S}^2$ and 
$\phi(S^1, S^2)\!=\!- \phi(S^2, S^1), \forall S^1, S^2 \in \mathbb{S}^1$; it is \emph{\textbf{transitive}} if there is  a monotonic \emph{rating function} $f$ such that $\phi(S^1, S^2)\!=\!f(S^1) - f(S^2), \forall S^1, S^2 \in \mathbb{S}^1$, meaning that performance on the game is the difference in ratings; it is  \emph{\textbf{non-transitive}} if $\phi$ satisfies $\sum_{S^2 \in \mathbb{S}^2} \phi(S^1, S^2) \!=\!0, \forall S^1 \in \mathbb{S}^1$, meaning that  winning against some strategies will be counterbalanced by losses against others;  the game has no consistent winner.  
Lastly, the \emph{\textbf{gamescape}} of a population of strategies \citep{balduzzi2019open} in a meta-game is defined as the convex hull of the payoff vectors of all policies in $\mathbb{S}$, written as:  
{{\begin{smalleralign}[\small] 
&\operatorname{Gamescape}\big({\mathbb{S}}\big) \nonumber \\ & := \Big\{  \sum_i \alpha_i \cdot \vm_i :  \boldsymbol{\alpha}\ge 0,  \boldsymbol{\alpha}^\top \mathbf{1} = 1,  \vm_i =\tM_{[i, :]}  \Big\}.	
\label{eq:gs}
\end{smalleralign}}}
\vspace{-15pt}
\subsection{Game Solvers}
\vspace{-5pt}
In solving NFGs, \emph{Fictitious play}  (FP)  \citep{brown1951iterative} describes the learning process where each player chooses a best response to their opponents' time-average strategies, and the resulting strategies guarantee to converge to the NE in two-player zero-sum, or   potential games. 
\emph{Generalised weakened fictitious play} (GWFP) \citep{leslie2006generalised} generalises FP by allowing for approximate best responses and perturbed average strategy updates. It is defined by:   
\vspace{-5pt}\begin{definition}[GWFP]
 GWFP is a process of  $\{\vpi_t\}_{t \geq 0}$ with $\vpi_t \in \prod_{i \in \mathcal{N}}\Delta_{\mathbb{S}^i}, $ following the below updating rule:  
{\begin{smalleralign}[\small]
    \vpi^i_{t+1} \in \big(1-\alpha_{t+1}\big)\vpi^i_t + \alpha_{t+1}\big(\mathbf{BR}^i_{\epsilon_t}(\vpi_t^{-i})+M^i_{t+1}\big). 
    \label{eq:gwfp}
\end{smalleralign}}
As  $t \rightarrow \infty$, $\alpha_t \rightarrow 0, \epsilon_t \rightarrow 0$ and $\sum_{t \geq 1}\alpha_n = \infty$. $\{M_t\}_{t\geq 1}$ is a sequence of perturbations that satisfies:  $ \forall T > 0$,
{{\begin{smalleralign}[\small]\lim _{t \rightarrow \infty} \sup _{k}\bigg\{ \Big\|\sum_{i=t}^{k-1} \alpha_{i+1} M_{i+1}\Big\| \text{ s.t. } \sum_{i=n}^{k-1} \alpha_{i} \leq T \bigg\}=0. \  \label{eq:gwfpc}\end{smalleralign}}}
GWFP recovers FP if  $\alpha_t = 1/t$, $\epsilon_t =0$ and $M_t=0, \forall t$. 
\vspace{-5pt}
\end{definition}




\begin{table}[t!]
\vspace{-10pt}
\caption{Variations of Different (Meta-)Game Solvers}
\vspace{-10pt}
\label{table:psros}
\begin{center}
   \resizebox{1.0\columnwidth}{!}{
  \begin{tabular}{p{74pt} c c p{50pt}}
  \toprule
   \textbf{Method} & \textbf{(Meta-)Policy} $\mathcal{S}$ & \textbf{Oracle} $\mathcal{O}$ & \textbf{Game type} \\  \midrule 
      Self-play \citep{fudenberg1998theory} & $[0,...,0,1]^N$ & $\mathbf{BR}(\cdot)$ & $\mathcal{N}$-player \ \  potential \\   \arrayrulecolor{black!30}\midrule 
GWFP \cite{leslie2006generalised}   & $\operatorname{UNIFORM}$ & $ \mathbf{BR}_\epsilon(\cdot)$ & 2-player zero-sum or potential \\ \arrayrulecolor{black!30}\midrule 
D.O. \cite{mcmahan2003planning}  & NE & $\mathbf{BR}(\cdot)$  & 2-player zero-sum \\ \arrayrulecolor{black!30}\midrule 
   PSRO$_N$ \cite{lanctot2017unified} & NE & $\mathbf{BR}_\epsilon(\cdot)$  & 2-player zero-sum \\ \arrayrulecolor{black!30}\midrule 
    PSRO$_{rN}$ \cite{balduzzi2019open} &  NE & Eq. (\ref{eq:rbr}) &  Symmetric zero-sum \\ \arrayrulecolor{black!30}\midrule 
    $\alpha$-PSRO \cite{muller2019generalized} & $\alpha$-Rank & Eq. (\ref{eq:pbr}) & $\mathcal{N}$-player general-sum \\ \arrayrulecolor{black}\midrule
    \textbf{Our Methods} & NE / $\alpha$-Rank & Eq. (\ref{eq:dpsro}) / (\ref{eq:dalpha}) & 2-player general-sum \\
   \arrayrulecolor{black} \bottomrule
  \end{tabular}}
\end{center}
\vspace{-15pt}
\end{table}
A general solver for open-ended (meta-)games involves an iterative  process of solving the equilibrium (meta-)policy first, and then based on the (meta-)policy,  finding a new better-performing policy to augment the existing population  (see the pseudocode in Appendix B. 
The (meta-)policy solver, denoted as $\mathcal{S}(\cdot)$,  computes a joint (meta-)policy profile  $\vpi$ based on the current payoff $\tM$ (or, $\mG$) where different solution concepts can be adopted (e.g., NE or $\alpha$-Rank). 
With $\vpi$, each agent then finds a new best-response policy, which is equivalent to solving a single-player optimisation problem against  opponents' (meta-)policies $\vpi^{-i}$. 
One can regard a best-response policy as given by an \emph{Oracle}, denoted by $\mathcal{O}$. In two-player zero-sum cases, an Oracle represents $\mathcal{O}^1(\vpi^2) =\{S^1: \sum_{S^2\in \mathbb{S}^2} \vpi^2(S^2) \cdot \phi (S^1, S^2) > 0 \}$. 
Generally,  Oracles  can  be  implemented through optimisation subroutines such as gradient-descent methods or RL algorithms. 
After a new policy is learned, the payoff table is expanded, and the missing entries will be filled by running new game simulations. 
The above  process loops over each player at every iteration, and it terminates if no  players can find new best-response policies  (i.e., Eq. (\ref{eq:nashconv})   reaches zero).


With  correct choices of (meta-)policy solver $\mathcal{S}$ and Oracle $\mathcal{O}$,  various types of (meta-)game solvers can be summarised in Table \ref{table:psros}. For example, it is trivial to see that GWFP is recovered when $\mathcal{S}=\operatorname{UNIFORM}(\cdot)$ and $\mathcal{O}^i=\mathbf{BR}_\epsilon^i(\cdot)$. 
Double Oracle (D.O.) and PSRO methods refer to the cases when the (meta-)solver computes NE. 
Notably, when $\mathcal{S} = \alpha$-Rank, \citet{muller2019generalized} showed  that the standard best response fails to converge to the SSCC of $\alpha$-Rank; instead,   they propose $\alpha$-PSRO where the Oracle is computed by the so-called \textit{Preference-based Best Response}  (PBR),  that is, 
{{
\begin{smalleralign}[\small]
\mathcal{O}^i\big(\vpi^{-i}\big)  \subseteq 
 \argmax_{\sigma \in \mathbb{S}^i} \mathbb{E}_{ \vpi^{-i}}\Big[ \mathds{1} \big[\tM^i(\sigma, S^{-i}) > \tM^i(S^{i}, S^{-i})   \big] \Big].
\label{eq:pbr}
\end{smalleralign}}}


\vspace{-15pt}
\subsection{Existing Diversity Measures}
\label{sec:diveristy}
\vspace{-5pt}
Promoting behavioural diversity can lead to  learning more effective strategies and achieving lower exploitability in performance. 
 The smooth FP method \citep{fudenberg1995consistency} incorporates the policy  entropy $\mathcal{H}(\pi)$ when finding the  best response to advocate diversity, written as  $\pi^i \in \mathbf{BR}_\epsilon^i(\vpi^{-i}) = \arg \max_{\pi \in \Delta_{S^i}}\big[\mG^i(\pi, \vpi^{-i}) + \tau \cdot \mathcal{H}(\pi)\big] $ where $\tau$ is a weighting hyper-parameter.   
In the case of $\tau \rightarrow 0$ as training goes on, smooth FP converges to the GWFP process almost surely \cite{leslie2006generalised}. 

Entropy measures the diversity of a policy in terms of its randomness; however, when it comes to solving open-ended (meta-)games, measuring diversity against peer models in the population  becomes critical. Towards this end, \emph{effective diversity} (ED) \citep{balduzzi2019open} is proposed to quantify the diversity for a population of policies $\mathbb{S}$ by 
\begin{smalleralign}[\small]
\hspace{-3pt}	\operatorname{ED}\big(\mathbb{S}\big) = {\vpi^{*}}^{\top}  \left\lfloor\tM\right\rfloor_{+} \vpi^{*}, \    \left\lfloor x\right\rfloor_{+}:= x \text{ if } x \ge 0 \text{ else } 0.   	\label{eq:ed}
	\vspace{-0pt}
\end{smalleralign}

$\tM$ is the meta-payoff table of $\mathbb{S}$, and $\vpi^*$ is the NE of $\tM$. 
The intuition of ED is that, 
using the Nash distribution ensures that the diversity is only related to the  best-responding models, and the \emph{rectifier} $\left\lfloor x\right\rfloor_{+}$ quantifies the number of variations of how those ``winner" models (those within the support of NE) beat each other. 
Under this design, if there is only one dominant policy in $\mathbb{S}$, then $\operatorname{ED}(\mathbb{S})=0$, thus no diversity.  
To promote ED in training,  a variation of PSRO -- PSRO$_{rN}$ -- is introduced, written as:   
\begin{smalleralign}[\small]
\hspace{-8pt}\mathcal{O}^1(\vpi^2) =\Big\{S^1: \sum_{S^2\in \mathbb{S}^2} \vpi^{2,*}(S^2) \cdot \lfloor\phi (S^1, S^2)\rfloor_{+} >0 \Big\}. 
\label{eq:rbr}
\end{smalleralign}

 In short, the ED in PSRO$_{rN}$  encourages players  to amplify its strengths and ignore its weaknesses in finding a new policy. On symmetric zero-sum games, if both players play their Nash strategy (this assumption will be removed  by our method), then  Eq. (\ref{eq:rbr}) guarantees to enlarge the gamescape. 

Nonetheless, focusing only on the winners can sometimes be problematic, since weak agents may still hold the promise of tackling niche tasks, and they can serve as stepping stones for discovering stronger policies later during training. 
For example, when training StarCraft AIs, overcoming agents' weaknesses was found to be more important than amplifying strengths \citep{vinyals2019grandmaster}, a completely opposite result to PSRO$_{rN}$.  
Another counter example  that fails PSRO$_{rN}$ is the RPS-X game  \citep{mcaleer2020pipeline}:   
\begin{smalleralign}
{\mG=\left[\begin{array}{cccc}0 & -1 & 1 & -{2}/{5} \\ 1 & 0 & -1 & -{2}/{5} \\ -1 & 1 & 0 & -{2}/{5} \\ {2}/{5} & {2}/{5} & {2}/{5} & 0\end{array}\right]}.
\label{eq:rpsx}
\end{smalleralign}
In RPS-X, if the initial strategy pool of  PSRO$_{rN}$  starts from either \{R\}, \{P\} or \{S\}, then the algorithm will terminate without exploring the fourth strategy because the best response to \{R,P,S\} is still in \{R,P,S\}; however, the  fourth strategy alone can still exploit the population of   \{R,P,S\} by getting a  positive payoff of $2/5$.   Also see in Appendix C 
how our method can tackle this problem. 




\vspace{-3pt}
\section{Our Methods}
\vspace{-3pt}
Instead of choosing between amplifying strengths or overcoming weaknesses,  we take an altogether different approach of modelling the behavioural diversity in games. 
Specifically, we introduce a new diversity measure based on a geometric interpretation of games modelled by a determinantal point process (DPP). Due to the space limit,  
all proofs in this section are provided in Appendix D. 

%

\vspace{-5pt}
\subsection{Determinantal Point Process}
\vspace{-5pt}
Originating in quantum physics for modelling repulsive Fermion particles \cite{macchi1977fermion,kulesza2012determinantal}, 
a DPP is a probabilistic framework that characterises how likely a subset of items is to be sampled from a ground set where diverse subsets are preferred. Formally, we have  

\vspace{-3pt}
\begin{definition}[DPP] 
\label{def:dpp}
For a ground set $\mathcal{Y}=\{1,2, ... , M\}$, a DPP defines  a probability measure $\mathbb{P}$ on  the power set of $\mathcal{Y}$ (i.e., $2^{\mathcal{Y}}$), such that, 
given  an  $M \times M$ positive semi-definite (PSD) kernel $\bm{\mathcal{L}}$ that measures the pairwise similarity for  items in $\mathcal{Y}$,  and 
let $\bm{Y}$ be a random subset drawn from the DPP,   the probability of sampling $ \forall Y \subseteq  \mathcal{Y} $  is written as      
\begin{smalleralign}[\small]
\operatorname{DPP}(\mathcal{L}):=  \mathbb{P}_{\bm{\mathcal{L}}}\big(\bm{Y} = Y\big) \propto \det	\big(\bm{\mathcal{L}}_Y\big) = \operatorname{Volume}^2\big(\{\bm{w}_i\}_{i \in Y}\big) \nonumber
\end{smalleralign}
where $\bm{\mathcal{L}}_Y :=  [\bm{\mathcal{L}}_{i,j}]_{i,j \in Y}$  denotes a submatrix of $\bm{\mathcal{L}}$  whose entries are indexed by the items included in $Y$. 
Given a  PSD kernel $\bm{\mathcal{L}}=\bm{\mathcal{W}}\bm{\mathcal{W}}^{\top}, \bm{\mathcal{W}} \in \mathbb{R}^{M \times P}, P \le M$, each row  $\bm{w}_i$ represents a $P$-dimensional feature vector of item $i \in \mathcal{Y}$, then the geometric meaning of $\det	(\bm{\mathcal{L}}_Y)$ is the squared volume of the parallelepiped spanned by the rows of $\bm{\mathcal{W}}$ that correspond to the sampled items in $Y$. 
\end{definition}
\vspace{-5pt}

A PSD matrix ensures  all principal minors of $\bm{\mathcal{L}}$ are non-negative (i.e., $\det	(\bm{\mathcal{L}}_Y) \ge 0,  \forall Y \subseteq  \mathcal{Y}$), which suffices to be a proper probability distribution. 
The normaliser of $\mathbb{P}_{\bm{\mathcal{L}}}(\bm{Y} = Y)$ can be computed by  $\sum_{Y \subseteq  \mathcal{Y}}\det	(\bm{\mathcal{L}}_Y) = \det	(\bm{\mathcal{L}} + \bm{I})$, where $\bm{I}$ is the $M\times M$ identity matrix.

The entries of $\bm{\mathcal{L}}$ are pairwise inner products between item  vectors.  The kernel $\bm{\mathcal{L}}$  can intuitively be thought of as representing dual effects -- the diagonal elements $\bm{\mathcal{L}}_{i,i}$ aim to capture the quality of item $i$, whereas the off-diagonal elements $\bm{\mathcal{L}}_{i,j}$ capture the similarity between the items $i$ and $j$.
A DPP models the \textbf{repulsive} connections among the items in a sampled subset. For example, in a two-item subset, since $\small \mathbb{P}_{\bm{\mathcal{L}}}\big(\{i, j\}\big) \propto \left|\begin{array}{ll}{\bm{\mathcal{L}}_{i,i}} & {\bm{\mathcal{L}}_{i,j}} \\ {\bm{\mathcal{L}}_{j,i}} & {\bm{\mathcal{L}}_{j,j}}\end{array}\right|=  \bm{\mathcal{L}}_{i,i}\bm{\mathcal{L}}_{j,j}-\bm{\mathcal{L}}_{i,j}\bm{\mathcal{L}}_{j,i}$, we know that if item $i$ and item $j$ are perfectly  similar such that $\bm{w}_i=\bm{w}_j$, and thus  $\bm{\mathcal{L}}_{i,j}=\sqrt{\bm{\mathcal{L}}_{i,i}\bm{\mathcal{L}}_{j,j}}$, then these two items will not co-occur, hence such a subset of $Y=\{i, j\}$ will be  sampled with probability zero. 

\vspace{-5pt}
\subsection{Expected Cardinality: A New Diversity Measure}
\vspace{-5pt}
\begin{figure}[t!]
\vspace{3pt}
\includegraphics[width=0.99\linewidth]{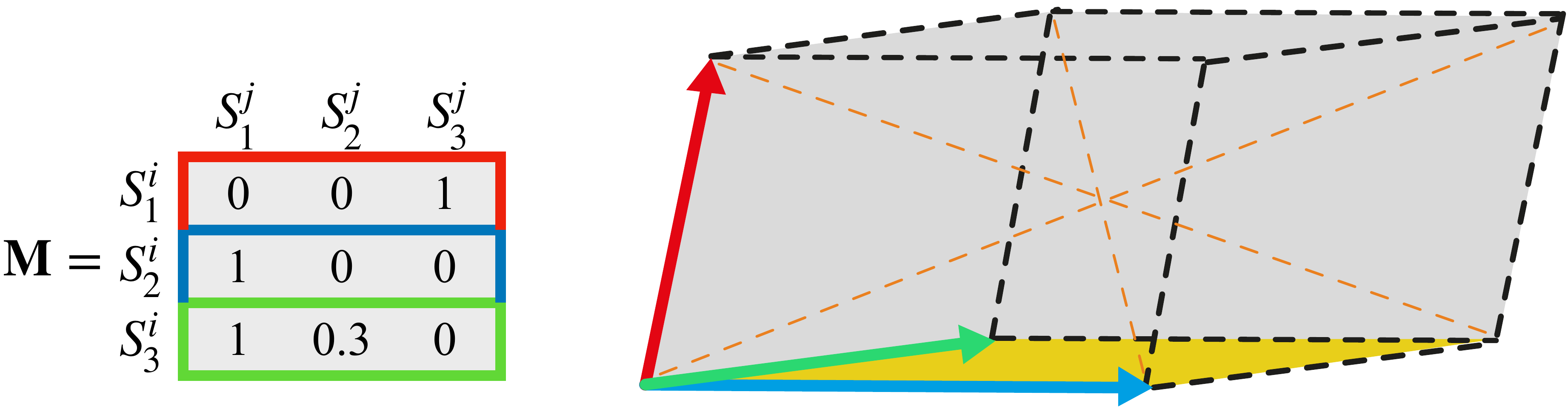}
\vspace{-0pt}
\caption{G-DPP. The squared volume of  the grey cube equals to $\det(\bm{\mathcal{L}}_{\{S^i_1, S^i_2, S^i_3\}})$.
 The probability of selecting $\{S^i_2, S^i_3\}$ from G-DPP (the yellow area) is smaller than that of selecting $\{S^i_1, S^i_2\}$ which has orthogonal payoff vectors.  The diversity in Eq. (\ref{eq:expcard}) of the population  $\{S^i_1\}$,$\{S^i_1, S^i_2\}$,$\{S^i_1, S^i_2, S^i_3\}$ are $0, 1, 1.21$.   }
\label{fig:gdpp}
\vspace{-5pt}
\end{figure}

Our target is to find a population of diverse  policies, with each of them performing differently from other policies due to their unique characteristics. Therefore, when modelling the behavioural diversity in games, 
 we can naturally use the payoff matrix to construct the DPP kernel so that the similarity between two policies depends on their performance  in terms of payoffs against different types of opponents.  
 \vspace{-3pt}
\begin{definition}[G-DPP, Fig. (\ref{fig:gdpp})]
A G-DPP for each player   is a DPP in which the ground set is the strategy  population $\mathcal{Y}=\mathbb{S}$, and the  DPP kernel $\bm{\mathcal{L}}$ is written by Eq. (\ref{eq:gdppl}), which is a  Gram matrix based on the payoff table $\tM$.  \begin{smalleralign} 
\bm{\mathcal{L}}_{\mathbb{S}} = \tM \tM^\top  	
\label{eq:gdppl}
\end{smalleralign}
\label{eq:gdpp}
\vspace{-20pt}
\end{definition}
For learning in open-ended games, we want to keep adding diverse policies to the population. 
This is equivalent to say, at each iteration,  if we take a random sample from the G-DPP that consists of all existing policies, we hope the cardinality of such a random sample is large (since policies with similar payoff vectors will be unlikely to co-occur!).   
In this sense, we can design a  diversity measure based on the expected cardinality of  random samples from a G-DPP, i.e.,  $\mathbb{E}_{\mathbf{Y} \sim \mathbb{P}_{\mathcal{L}_\mathbb{S}}}\big[|\mathbf{Y}|\big] $.
By the following proposition, we show that computing such a diversity measure is tractable. 
\begin{proposition}[G-DPP Diversity Metric] \label{lemma:expcard}
	The diversity metric, defined as the expected cardinality of a G-DPP, can be computed in $\mathcal{O}(|\mathbb{S}|^3)$ time by the following equation:  
	\begin{smalleralign}[\small]
  \operatorname{Diversity}\big(\mathbb{S}\big)=  \mathbb{E}_{\mathbf{Y} \sim \mathbb{P}_{\mathcal{L}_\mathbb{S}}}\big[|\mathbf{Y}|\big] &= \Tr  \big(\mathbf{I} - (\mathcal{L}_\mathbb{S} + \mathbf{I})^{-1}\big).   
    \label{eq:expcard}
\end{smalleralign}
\end{proposition}
A nice property of our diversity measure is that it is well-defined even in the case when  $\mathcal{Y}$ has duplicated policies, as dealing with redundant policies turns out to be a critical challenge for game evaluation \citep{balduzzi2018re}. In fact, redundancy also prevents us from directly using $\operatorname{det}\big(\mathcal{L_\mathbb{S}}\big)$ as the diversity measure because the determinant value becomes zero with duplicated entries. 

\textbf{Expected Cardinality \emph{vs.} Matrix Rank.}
There is a fundamental difference between using expected cardinality and using the rank of a payoff matrix as the diversity measure. The matrix rank is the maximal number of linearly independent columns, though it can measure the \emph{difference} between the columns, it cannot model the \emph{diversity}.  
 For example, in RPS, a strategy of [$99\%$ Rock, $1\%$ Scissor] and a strategy of [$98\%$ Rock, $2\%$ Scissor] are different but they are not diverse as they both favour playing Rock. 
If one strategy is added  into the population whilst the other already exists,  the rank of the payoff matrix will increase by one, but the increment on expected cardinality is minor. In Fig.  (\ref{fig:gdpp}), adding the green strategy only contributes to the expected cardinality by  $0.21$. 
This property is  particularly important for learning in games, in the sense that finding a \emph{diverse} policy is often harder than finding just a  \emph{different} policy.
To summarise, we show  the following proposition. 
\vspace{-2pt}
\begin{proposition}[Maximum Diversity]
The diversity of a population  $\mathbb{S}$ is bounded by  
    $\operatorname{Diversity}\big(\mathbb{S}\big) \leq \operatorname{rank}(\tM)$, and if $\tM$ is normalised (i.e., $||\tM_{[i,:]}|| = 1, \forall i$),  
    we have $\operatorname{Diversity}\big(\mathbb{S}\big) \leq \operatorname{rank}(\tM)/2$. In both cases, maximal diversity is reached if and only if  $\tM$ is orthogonal.  
    \label{prop:bound}
\end{proposition}
\textbf{Expected Cardinality \emph{vs.} Effective Diversity.}
We also argue that the  principles that underpin Eq. (\ref{eq:ed})  and Eq. (\ref{eq:expcard}) are different. Here we illustrate from the perspective of matrix norm. 
Notably,  maximising the effective diversity in Eq. (\ref{eq:ed})
is equivalent to maximising a matrix norm, in the sense that $\operatorname{ED}(\mathbb{S}) = \frac{1}{2}\left\|\vpi^* \odot {\tM} \odot \vpi^* \right\|_{1,1}$ where $\odot$ is the Hadamard product and $\|\mathbf{A}\|_{1,1}:=\sum_{i j}\left|a_{i j}\right|$.
In comparison, the  proposition below shows that maximising our diversity measure in Eq. (\ref{eq:expcard}) will also maximise the Frobenius norm of $\tM$.
\begin{proposition}[Diversity vs. Matrix Norm] Maximising the diversity in Eq. (\ref{eq:expcard}) also maximises the Frobenius norm of $\|\tM\|_F$, but NOT vice versa. 
\label{prop:fnorm}
\vspace{-5pt}
\end{proposition}
Geometrically, for a given matrix $\tM$, considering the box which is the image of a unit cube (in the 3D case) that is stretched by $\tM$, the Frobenius norm represents the sum of lengths of all diagonals in that box regardless of their directions (the orange lines in Fig. (\ref{fig:gdpp})). Therefore, whilst 
the $\|\cdot\|_{1,1}$ norm reflects the idea that $\operatorname{ED}(\mathbb{S})$ in Eq. (\ref{eq:ed})  accounts for the winners within the Nash support only, the Frobenius norm, on the contrary, considers all strategies' contribution to diversity.  
We show later that this results in significant performance improvements over PSRO$_{rN}$. 

Notably, it is worth highlighting that the opposite direction of Proposition \ref{prop:fnorm} is not correct, that is,  maximising $\|\tM\|_F$ will \textbf{NOT} necessarily lead to a large diversity. 
A counter-example in Fig. (\ref{fig:gdpp}) is that, if one of the orange lines is long but the rest are short, though the Frobenius norm is large, the expected cardinality is still  small. Thus, the diversity metric in Eq. (\ref{eq:expcard}) cannot simply be replaced by $\|\tM\|_F$. We also provide empirical evidence in Appendix F. 

\vspace{-5pt}
\subsection{Diverse Fictitious Play}
\vspace{-5pt}
With the newly proposed diversity measure of Eq. (\ref{eq:expcard}), we can now design diversity-aware learning algorithms. We start by extending the classical FP to a diverse version such that at each iteration, the player not only considers a best response, but also considers how this new strategy can help enrich the existing strategy pool after the update. 
Formally, our \emph{diverse FP} method maintains the same update rule as Eq. (\ref{eq:gwfp}), but with the best response changing into 
\begin{smalleralign}[\small]
&\mathbf{BR}_\epsilon^i(\vpi^{-i}) \nonumber \\ & = \argmax_{\vpi \in \Delta_{\mathbb{S}^i}}\Big[\mG^i \big(\vpi, \vpi^{-i}\big) + \tau \cdot   \operatorname{Diversity}\big(\mathbb{S}^i \cup \{\vpi \} \big) \Big]
\label{eq:dbr}
\end{smalleralign}
where $\tau$ is a tunable constant. 
A nice property of diverse FP is that the expected cardinality is guaranteed to be a strictly concave function; therefore, Eq. (\ref{eq:dbr}) has a unique solution at each iteration. We have the following proposition:

\begin{proposition}[Uniqueness of Diverse Best Response]
	Eq. (\ref{eq:expcard}) is a strictly concave function. The resulting best response in Eq. (\ref{eq:dbr})  has a unique solution.
	\label{prop:concave}
\end{proposition}
Intuitively, the diverse FP process will almost surely converge to a GWFP process  as long as $\tau \rightarrow 0$ and thus will enjoy the same convergence guarantees as GWFP (i.e., to a NE in two-player zero-sum or potential games).
However, in order to prove such connection rigorously, we need to show the sequence of expected changes in strategy, which is induced by finding a strategy that maximises Eq. (\ref{eq:dbr}) at each iteration, is actually a uniformly bounded martingale sequence that satisfies Eq. (\ref{eq:gwfpc}). We show the below theorem: 
\begin{theorem}[Convergence of Diverse FP]
The perturbation sequence induced by diverse FP process is a uniformly bounded martingale difference sequence; therefore, diverse FP shares the same convergence property as GWFP. 
 \label{theorem:convergence}
 \vspace{-0pt}
\end{theorem}

\vspace{-5pt}
\subsection{Diverse Policy-Space Oracle}
\vspace{-5pt}
When solving NFGs, the total number of pure strategies is known and thus a best response  in Eq. (\ref{eq:dbr}) can be computed through a direct search, and the uniqueness of the solution is guaranteed by Proposition \ref{prop:concave}.  When it comes to solving open-ended (meta-)games, the total number of policies is unknown and often infinitely many. Therefore, a best response has to be computed through optimisation subroutines such as gradient-based methods or RL algorithms. 
Here we extend our diversity measure to the policy space and develop diversity-aware solvers for open-ended (meta-)games. 

In solving open-ended games, at the $t$-th  iteration, the algorithm maintains a population of policies  $\mathbb{S}_t^i$  learned so far by player $i$. 
Our goal here is to design an Oracle to train a new strategy $S_\vtheta$, parameterised by $\vtheta\in \mathbb{R}^d$ (e.g., a deep neural net), which both maximises player $i$'s  payoff and  is diverse from  all strategies in  $\mathbb{S}_t^i$. 
Therefore, we define the ground set of the G-DPP at iteration $t$ to be the union of the existing  $\mathbb{S}_t^i$ and the new model to add:  
$
    \mathcal{Y}_t = \mathbb{S}_t^i \cup \big\{S_\vtheta \big\}. \nonumber
$

With the ground set at each iteration, we can compute the diversity measure by Eq. (\ref{eq:expcard}).  
Subsequently, the objective of an Oracle can be written as
{\begin{smalleralign}[\small]
 \mathcal{O}^1(\vpi^2) = \argmax_{\vtheta \in \mathbb{R}^d}  \sum_{S^2  \in \mathbb{S}^2} & \vpi^2\big(S^2 \big) \cdot \phi \big(S_\vtheta, S^2\big) \label{eq:dpsro} \\ & + \tau \cdot \operatorname{Diversity}\Big(\mathbb{S}^1 \cup \big\{S_\vtheta \big\} \Big) \nonumber 
\end{smalleralign}}
where $\vpi^2(\cdot)$ is the policy of the player two; depending on the game solvers, it can be NE, $\operatorname{UNIFORM}$, etc. 

  
  Based on Eq. (\ref{eq:dpsro}), we can tell that the diversity of policies during training  comes from two aspects.  
 The obvious aspect is from the expected cardinality of the G-DPP that forces agents to find diverse  policies.   
  The less obvious aspect is from how the opponents are treated. Although the (meta-)policy of player $2$ is determined by $\vpi^2(\cdot)$, the learning player will have to focus on exploiting certain  aspects of $\vpi^2(\cdot)$ in order to acquire diversity. This is similar in manner to selecting a diverse set of opponents. 
Theoretically, we are able to show that our diversity-aware Oracle can strictly enlarge the gamescape. 
Unlike PSRO$_{rN}$ (see Proposition 6  in \citet{balduzzi2019open}), we do \textbf{NOT} need to assume the opponents are playing NE before reaching the result below.   
\begin{proposition}[Gamescape Enlargement] 
Adding a new best-response policy $S_\vtheta$ via Eq. (\ref{eq:dpsro}) strictly enlarges the gamescape. Formally, we have 
\begin{smalleralign}[\small]
\operatorname{Gamescape}\big({\mathbb{S}}\big)  \subsetneq \operatorname{Gamescape}\Big({\mathbb{S}\cup \big\{S_\vtheta \big\}}\Big). \nonumber
\end{smalleralign}
\end{proposition}
\vspace{-5pt}

  
%

\textbf{Implementation of Oracles.}
When  the game engine $\phi$ is differentiable, we can directly apply gradient-based methods to solve  Eq. (\ref{eq:dpsro}).  In general, many real-world games are black-box, thus
we have to seek for gradient-free solutions or model-free RL algorithms. To tackle this, we provide zero-order Oracle and RL-based Oracle as approximation solutions to Eq. (\ref{eq:dpsro}), and list their pseudocode and time complexity in Appendix H. 

\begin{figure}[t!]
\vspace{-0pt}
\includegraphics[width=0.95\linewidth]{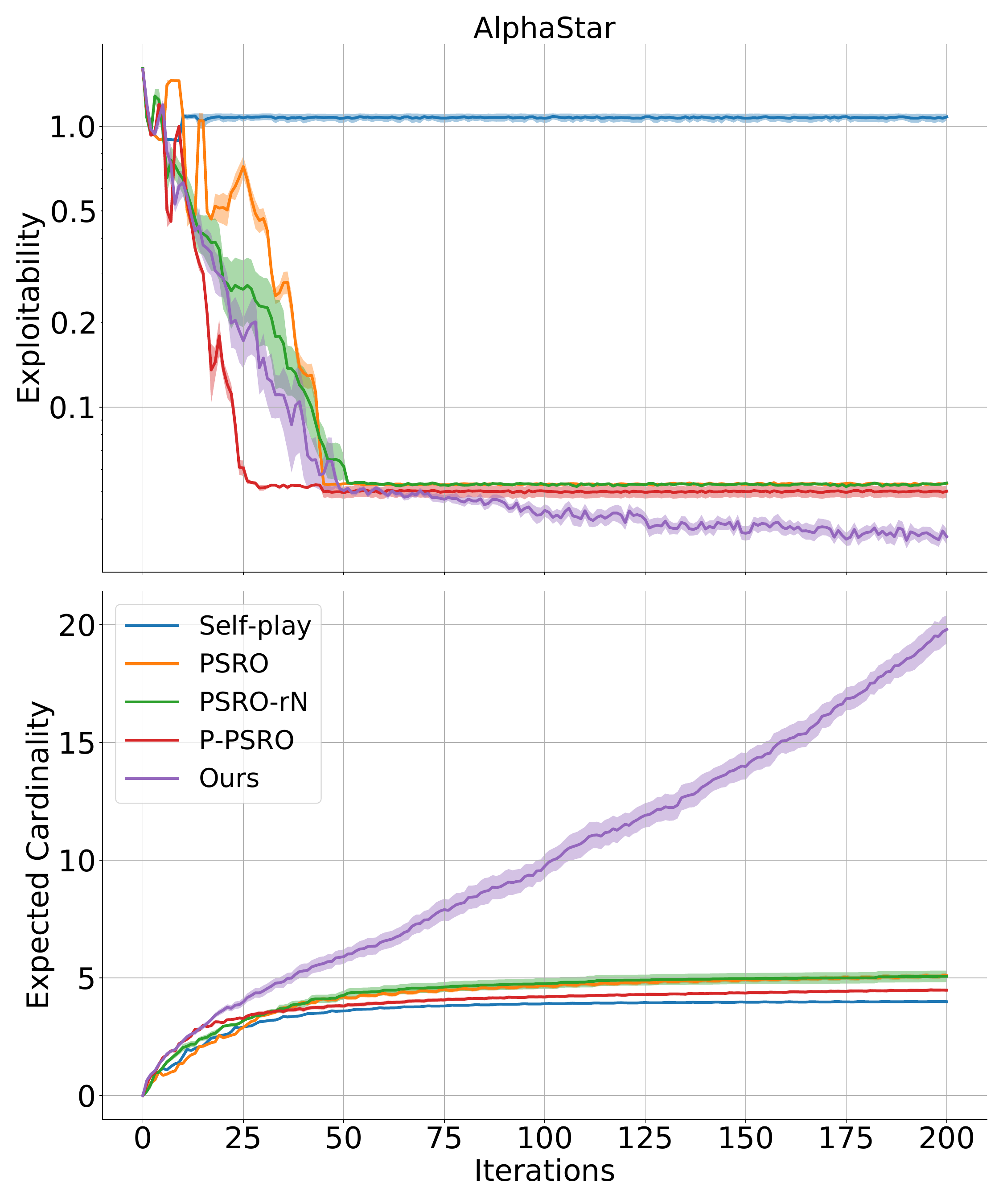}
\vspace{-10pt}
\caption{Exploitability and diversity \emph{vs.} training iterations (number of times a solution concept is computed) on the AlphaStar meta-game (size $888 \times 888$). Our method achieves the lowest exploitability by finding a diverse population of $50$ policies. }
\vspace{-15pt}
\label{fig:games_skill}
\end{figure}

\vspace{-5pt}
\subsection{Diverse Oracle for $\alpha$-Rank}
\vspace{-5pt}
We also develop diverse Oracles that suit $\alpha$-Rank. 
Note that $\alpha$-Rank is a replacement solution concept for NE on $\mathcal{N}$-player general-sum games;  therefore, the goal of learning is  finding all  SSCCs on the response graph. 
Since the standard best response does not have convergence guarantees, we introduce a diversity-aware extension based on $\alpha$-PSRO \cite{muller2019generalized} whose Oracle is written in Eq. (\ref{eq:pbr}). 
Specifically, we  adopt the quality-diversity decomposition of DPP \cite{affandi2014learning} to unify Eq. (\ref{eq:pbr}) and Eq. (\ref{eq:expcard}). 
Given $\bm{\mathcal{L}}=\bm{\mathcal{W}}\bm{\mathcal{W}}^\top$, we can rewrite the $i$-th row of $\bm{\mathcal{W}}$ to be the product of a quality term $q_i \in \mathbb{R}^+$ and a diversity feature $\boldsymbol{w}_i \in \mathbb{R}^P$, thus 
    $\mathcal{L}_{ij} = q_i \boldsymbol{w}_i \boldsymbol{w}_j^\top q_j $.  
We design the quality term to be the exponent of the PBR value in Eq. (\ref{eq:pbr}), and the diversity feature follows G-DPP in Eq. (\ref{eq:gdppl}), that is,  
\begin{smalleralign}[\small]
    q_i = \exp \Big(  \mathbb{E}_{ \vpi^{-i}}\big[\mathds{1} [\tM^i(\sigma, S^{-i}) > \tM^i(S^{i}, S^{-i})   ] \big]\Big), \boldsymbol{w}_i = \dfrac{\tM_{[i,:]}} { \|\tM \|_F}. \nonumber 
\end{smalleralign}
The resulting diversity-aware Oracle that suits $\alpha$-Rank is:  
%
%
%
{
\begin{smalleralign}[\small]
\mathcal{O}_t^i(\vpi^{-i})  = \argmax_{\vpi \in \Delta_{\mathbb{S}^i}} \Tr \Big(\mathbf{I} - \big(\bm{\mathcal{L}}_{\mathbb{S}^i_t \cup \{\vpi \}} + \mathbf{I}\big)^{-1}\Big).
\label{eq:dalpha}
\end{smalleralign}}
\vspace{-15pt}

The  following theorem shows the convergence result of our diverse $\alpha$-PSRO to SSCC on two-player symmetric NFGs. 
\begin{theorem}[Convergence of Diverse $\alpha$-PSRO] 
\label{thm:alpha}
Diverse $\alpha$-PSRO with the Oracle of Eq. (\ref{eq:dalpha}) converges to the sub-cycle of the unique SSCC in the two-player symmetric games.
\end{theorem}
%




%

\begin{figure*}[t!]
\vspace{-5pt}
\centering
\includegraphics[width=1.0\linewidth]{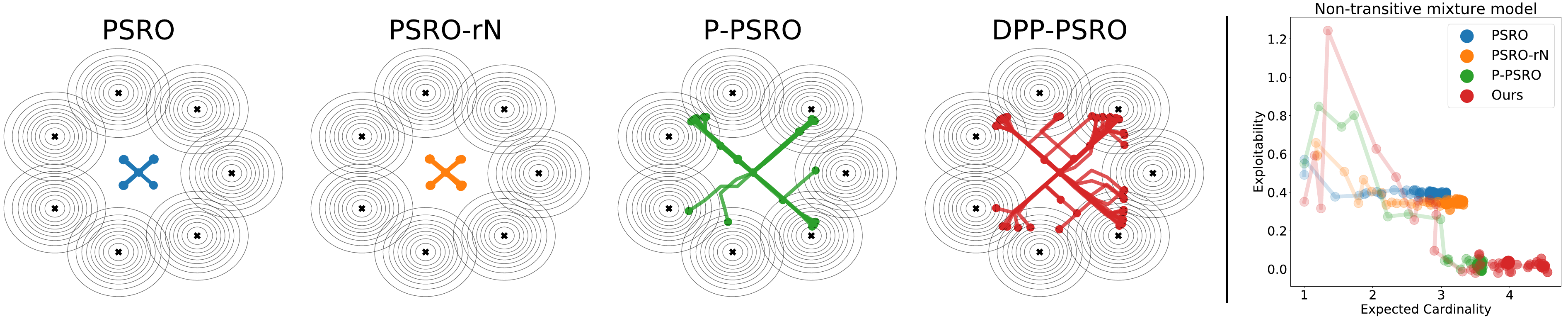}
\vspace{-25pt}
\caption{Non-transitive mixture model. Exploration trajectories during training and  Performance vs. Diversity comparisons. }
\label{fig:2d-RPS}
\end{figure*}

\begin{figure*}[t!]
\vspace{-5pt}
\centering
\includegraphics[width=0.95\linewidth]{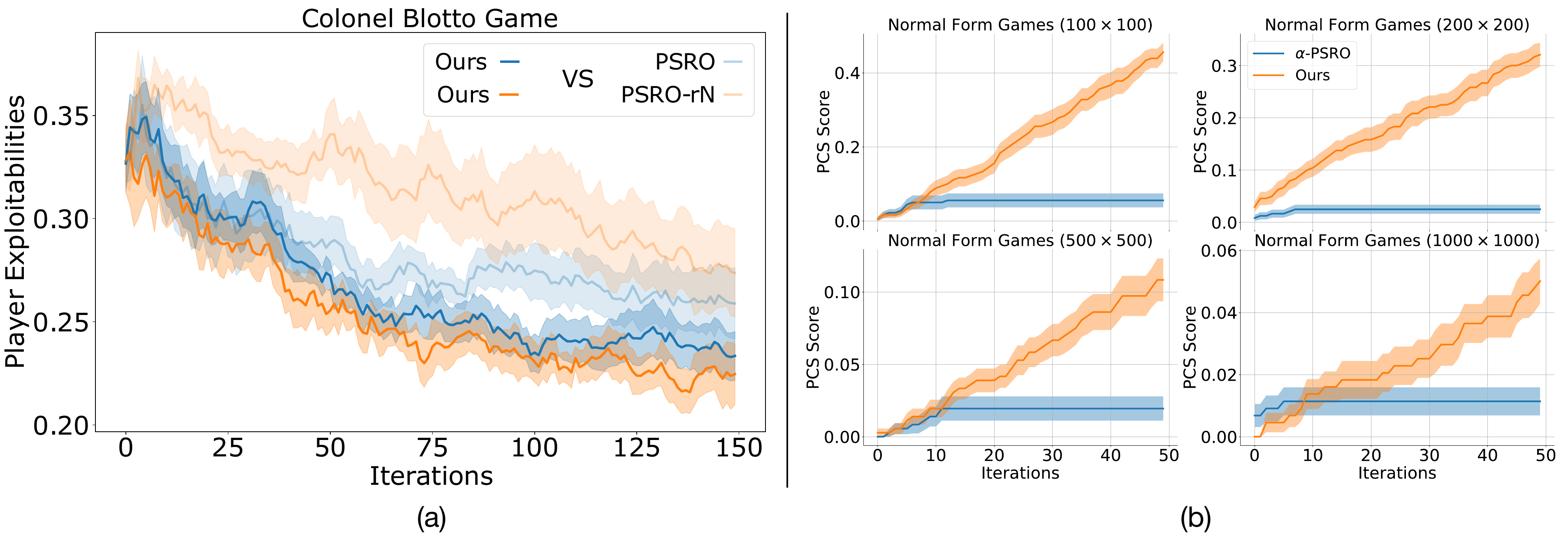}
\vspace{-10pt}
\caption{a) Performance of our diverse PSRO \emph{vs.} PSRO, diverse PSRO \emph{vs.} PSRO$_{rN}$ on the Blotto Game, b) PCS-Score comparison of our diverse $\alpha$-PSRO \emph{vs.} $\alpha$-PSRO on NFGs with variable sizes.}
\label{fig:blotto}
\vspace{-10pt}
\end{figure*}

\vspace{-10pt}
\section{Experiments \& Results}
\vspace{-3pt}
We compare our diversity-aware solvers with state-of-the-art game solvers  including self-play,  PSRO \citep{lanctot2017unified}, Pipeline-PSRO \cite{mcaleer2020pipeline}, rectified PSRO \citep{balduzzi2019open}, and $\alpha$-PSRO \citep{muller2019generalized}. 
We investigate the performance of these algorithms on both NFGs and open-ended games. 
Our selected games  involve both transitive and non-transitive dynamics. 
If an algorithm fails to discover a diverse set of policies, it will be trapped in some local strategy cycles that are easily exploitable (e.g., recall the  illustrative   example of the RPS-X game  in Section \ref{sec:diveristy}, and see how our method can tackle this game  in Appendix C. 
Therefore, we focus on the evaluation metrics of exploitability in Eq. (\ref{eq:nashconv}) and how extensively the gamescapes are explored. We note that the confidence intervals represented in Figs. (\ref{fig:games_skill}, \ref{fig:blotto}a, \ref{fig:blotto}b) represent the standard deviation in the exploitability at each iteration over multiple seeds, where the number of seeds is reported in Appendix G.
One exception is the comparison between $\alpha$-PSRO and diverse $\alpha$-PSRO, since the solution concept is $\alpha$-Rank, instead of exploitability that measures distance to a NE, we apply the metric of 
PCS-score  \cite{muller2019generalized} -- the number of SSCC that has been found -- for fair comparison.
We provide an exhaustive list of hyper-parameter and reward settings  in Appendix G. 


\textbf{Real-World Meta-Games.} We test our methods on the meta-games that are generated during the process of solving 28 real-world games \cite{czarnecki2020real}, including AlphaStar and AlphaGO.
In Fig. (\ref{fig:games_skill}), we report the results over  the AlphaStar game that contained the meta-payoffs for $888$ RL policies, and report the results of the other 27 games in Appendix E. 
We used Algorithm 2 in Appendix H where agents are defined at the metagame level and correspond to mixed strategies of the underlying game. The results show that our diverse-PSRO method will, at worst, perform
as well as existing PSRO baselines, but in many cases
(e.g., Fig \ref{fig:games_skill},\ref{fig:2d-RPS},\ref{fig:blotto}) will outperform in terms of exploitability,
and will always outperform in terms of diversity. In particular,
we believe that the performance advantage comes
from the fact that without accounting for behavioural diversity,
PSRO baselines tend to enter into a cyclic phase where
repetitive strategies already in the population are found,
whereas our diversifying measure can help discover novel
strategies that consequently lead to lower exploitability. While many of the baselines have saturated in finding diverse strategies, our method keeps finding novel effective strategies which leads to a near zero exploitability in almost all $28$ games. 
In AlphaStar, our method achieves the best performance by only using less than $50$ out of $888$ RL policies, and with the population size growing, the exploitability keeps approaching zero while other methods saturate.



\textbf{Non-Transitive Mixture Model.}
This game consists of seven equally-distanced Gaussian humps on the 2D plane. Each strategy  corresponds to a point on the 2D plane, which, equivalently, represents the weights that each player puts on the humps, measured by the likelihood of that point in each  Gaussian distribution. The payoff of the game that includes both non-transitive and transitive components is given by: 
 \begin{smalleralign}[\scriptsize]
 \vpi^{1,\top} \tiny \left[\begin{array}{ccccccc}
0 & 1 & 1 & 1 & -1 & -1 & -1 \\ 
-1 & 0 & 1 & 1 & 1 & -1 & -1 \\ 
-1 & -1 & 0 & 1 & 1 & 1 & -1 \\ 
-1 & -1 & -1 & 0 & 1 & 1 & 1 \\ 
1 & -1 & -1 & -1 & 0 & 1 & 1 \\ 
1 & 1 & -1 & -1 & -1 & 0 & 1 \\ 
1 & 1 & 1 & -1 & -1 & -1 & 0
\end{array}\right] \vpi^{2} + \frac{1}{2}\sum_{k=1}^7 (\vpi^{1}_k-\vpi^{2}_k). \nonumber \end{smalleralign}
Since there are infinite number of points on the 2D plane, this game is open-ended.  
A winning player must learn to stay close to the Gaussian centroids  whilst also exploring all seven Gaussians to avoid being exploited. 
In Fig. (\ref{fig:2d-RPS}), we show the exploration trajectories for different algorithms along with the plot of exploitability \emph{vs.} diversity. 
Results suggest that both PSRO and PSRO$_{rN}$ fail to complete the task; we believe it is due to the same reason as RPS-X where strategy cycling occurs. In contrast,  
DPP-PSRO  solves the task almost perfectly, reaching zero exploitability, by generating a population of diverse and effective strategies.

%


\textbf{Colonel Blotto.} Blotto is a classical resource allocation game that is widely analysed for election campaigns \citep{roberson2006colonel}. In this game, two players have a budget of coins which they simultaneously distribute over a fixed number of areas. An area is won by the player who puts the most coins, and the player that wins the most areas wins the game. 
We report the results on the game with $3$ areas and $10$ coins over $10$ games. 
We test how a diverse PSRO player performs in terms of exploitability against a PSRO and a PSRO$_{rN}$ player, respectively. 
Fig. (4a) shows that our method (dark colours) consistently achieves a lower exploitability than the opponent player of either PSRO or PSRO$_{rN}$ (light colours). 

\textbf{Diverse $\alpha$-PSRO.}
As the PBR in Eq. (\ref{eq:pbr}) requires looping through all strategies in $S^i$,  
we test our  method on randomly generated zero-sum NFGs with varying dimensions. 
We do not employ the \textit{novelty-bound} suggested in \citet{muller2019generalized} to illustrate how the original $\alpha$-PSRO displays strong cyclic behaviour, which stops it from finding even a few underlying SSCC elements. Results in Fig. (\ref{fig:blotto}b) suggest that our diverse $\alpha$-PSRO  can effectively prevent the learner from exploring the same strategic cycles during training; it is therefore able to find more SSCCs of $\alpha$-Rank, and  outperform   $\alpha$-PSRO on the PCS-score. 


\vspace{-5pt}
\section{Conclusion}
We offer a geometric interpretation of behavioural diversity for learning in games by introducing a new diversity measure built upon the expected cardinality of a DPP. 
Based on the diversity metric, we propose general solvers for normal-form games and open-ended (meta-)games. We prove the convergence of our methods to NE and $\alpha$-Rank in two-player games, and show  theoretical guarantees of expanding the gamescapes.  
On tens of games, our methods achieve lower  exploitability than PSRO variants by finding both effective and diverse strategies.

\bibliography{mybibfile}
\bibliographystyle{icml2021}

\clearpage


\section*{Supplementary material}

# Supplementary Material for
# Modelling Behavioural Diversity for Learning in Open-Ended Games

**Contents**

## A. A Review of $\alpha$-Rank

$\alpha$-Rank (Omidshafiei et al., 2019) is a new type of solution concept that is built on the *response graph* of a game. In particular, it tries to find the so-called *sink strongly-connected components (SSCC)* nodes on the response graph that have only incoming edges but no outgoing edges. $\alpha$-Rank serves as a promising replacement for NE; the key associated benefits are its uniqueness, and its polynomial-time solvability in multi-player general-sum games.

On the response graph, each joint pure-strategy profile is a node, and a directed edge points from node $\sigma \in \mathbb{S}$ to node $S \in \mathbb{S}$ if 1) $\sigma$ and $S$ differ in only one single player's strategy, and 2) that deviating player, denoted by $i$, benefits from deviating from $S$ to $\sigma$ such that $\boldsymbol{G}^i(\sigma) > \boldsymbol{G}^i(S)$. We are interested in the so-called *sink strongly-connected components (SSCC)* nodes that have only incoming edges but no outgoing edges on the response graph. To find those SSCC nodes, $\alpha$-Rank constructs a random walk along the directed response graph, which can be equivalently described by a Markov chain, with the transition probability matrix $\boldsymbol{C}$ being:

$$\boldsymbol{C}_{S,\sigma} = \begin{cases} \eta \frac{1-\exp\left(-\alpha\left(\boldsymbol{G}^k(\sigma)-\boldsymbol{G}^k(s)\right)\right)}{1-\exp\left(-\alpha m\left(\boldsymbol{G}^k(\sigma)-\boldsymbol{G}^k(s)\right)\right)} & \text{if } \boldsymbol{G}^k(\sigma) \neq \boldsymbol{G}^k(S) \\ \frac{\eta}{m} & \text{otherwise} \end{cases}$$

$$\boldsymbol{C}_{S,S} = 1 - \sum_{i \in \mathcal{N}} \boldsymbol{C}_{S,\sigma}$$

In the above Equation, $\eta = (\sum_{i \in \mathcal{N}}(|S^i| - 1))^{-1}$, $m \in \mathbb{N}$ and $\alpha > 0$ are three constants. Large $\alpha$ ensures the Markov chain is irreducible, and thus guarantees the existence and uniqueness of the $\alpha$-Rank solution, which is the resulting unique stationary distribution $\boldsymbol{\pi}$ of the Markov chain, $\boldsymbol{C}^\top \boldsymbol{\pi} = \boldsymbol{\pi}$. The probability mass of each joint strategy in $\boldsymbol{\pi}$ can be interpreted as the longevity of that strategy during an evolution process.

## B. Algorithm for Open-Ended (Meta-)Game Solvers

---
**Algorithm 1** A General Solver for Open-Ended (Meta-)Games

---
1: **Initialise:** the "high-level" policy set $\mathbb{S} = \prod_{i \in \mathcal{N}} \mathbb{S}^i$, the meta-game payoff $\mathbf{M}, \forall S \in \mathbb{S}$, and meta-policy $\boldsymbol{\pi}^i =$ UNIFORM($\mathbb{S}^i$).
2: **for** iteration $t \in \{1, 2, ...\}$ **do**:
3:    **for** each player $i \in \mathcal{N}$ **do**:
4:       Compute the meta-policy $\boldsymbol{\pi}_t$ by meta-game solver $\mathcal{S}(\mathbf{M}_t)$.
5:       Find a new policy against others by Oracle: $S_t^i = \mathcal{O}^i(\boldsymbol{\pi}_t^{-i})$.
6:       Expand $\mathbb{S}_{t+1}^i \leftarrow \mathbb{S}_t^i \cup \{S_t^i\}$ and update meta-payoff $\mathbf{M}_{t+1}$.
7:   **terminate if:** $\mathbb{S}_{t+1}^i = \mathbb{S}_t^i, \forall i \in \mathcal{N}$.
8: **Return:** $\boldsymbol{\pi}$ and $\mathbb{S}$.

---

# C. The Effectiveness of Our Method on the RPS-X Game

We show via a worked example why our method will not fail to find the final strategy X in the RPS-X game, whereas current diversifying methods do fail.[1] Assume that we are in the state of the strategy space being $\{R, P, S\}$, we know that $\text{PSRO}_{rn}$ from this position will fail as the best response strategy is still within $\{R, P, S\}$ (McAleer et al., 2020). Instead, we show that if we follow our method and select the strategy with the largest expected cardinality as the best response, that strategy $\{X\}$ will be selected to be added to the population. There are four cases to consider, namely those of adding all four strategies to the current population of strategies:

1. Strategy $\{R\}$ is added to the population and we have the full strategy set $\{R, P, S, R\}$, which induces the following meta-game:

$$\mathbf{M} = \begin{bmatrix} 0 & -1 & 1 & 0 \\ 1 & 0 & -1 & 1 \\ -1 & 1 & 0 & -1 \\ 0 & -1 & 1 & 0 \end{bmatrix}.$$

and the following kernel matrix:

$$\mathcal{L} = \mathbf{M}\mathbf{M}^T = \begin{bmatrix} 2 & -1 & -1 & 2 \\ -1 & 3 & -2 & -1 \\ -1 & -2 & 3 & -1 \\ 2 & -1 & -1 & 2 \end{bmatrix}.$$

which we can use to solve for the expected cardinality:

$$\text{Tr}\left(\mathbf{I} - (\mathcal{L} + \mathbf{I})^{-1}\right) = 1.6667$$

2. Strategy $\{P\}$ is added to the population and we have the full strategy set $\{R, P, S, P\}$. The result is the same as that of adding $\{R\}$ to the population and the expected cardinality is 1.6667.

3. Strategy $\{S\}$ is added to the population and we have the full strategy set $\{R, P, S, S\}$. The result is the same as that of adding $\{R\}$ to the population and the expected cardinality is 1.6667.

4. Strategy $\{X\}$ is added to the population and we have the full strategy set $\{R, P, S, X\}$, which induces the following meta-game:

$$\mathbf{M} = \begin{bmatrix} 0 & -1 & 1 & -2/5 \\ 1 & 0 & -1 & -2/5 \\ -1 & 1 & 0 & -2/5 \\ 2/5 & 2/5 & 2/5 & 0 \end{bmatrix}.$$

and the following kernel matrix:

$$\mathcal{L} = \mathbf{M}\mathbf{M}^T = \begin{bmatrix} 2.16 & -0.84 & -0.84 & 0 \\ -0.84 & 2.16 & -0.84 & 0 \\ -0.84 & -0.84 & 2.16 & 0 \\ 0 & 0 & 0 & 0.48 \end{bmatrix}.$$

which we can use to solve for the expected cardinality:

$$\text{Tr}\left(\mathbf{I} - (\mathcal{L} + \mathbf{I})^{-1}\right) = 2.1486$$

which, due to inducing the largest expected cardinality out of all of the possible options to be added to the population, strategy $\{X\}$ would be added.

---

[1]Notably, vanilla PSRO will solve this game in this setting as it will just collapse to the Double Oracle method.

# D. Full Proofs

## D.1. Proposition 4 [G-DPP Diversity Metric]

**Proposition 4** (**G-DPP Diversity Metric**). *The diversity metric, defined as the expected cardinality of a G-DPP, can be computed in $\mathcal{O}(|\mathbb{S}|^3)$ time by the following equation:*

$$\text{Diversity}(\mathbb{S}) = \mathbb{E}_{\mathbf{Y} \sim \mathbb{P}_{\mathcal{L}_{\mathbb{S}}}}[|\mathbf{Y}|] = \text{Tr}\left(\mathbf{I} - (\mathcal{L}_{\mathbb{S}} + \mathbf{I})^{-1}\right).$$

**Proof.** We can calculate the expected cardinality of a DPP sample by an exponential sum over all subsets of $\mathbb{S}$ as follows (Gillenwater et al., 2018):

$$\mathbb{E}_{\mathbf{Y} \sim \mathbb{P}_{\mathcal{L}_{\mathbb{S}}}}[|\mathbf{Y}|] = \sum_{\mathbf{Y} \subseteq \mathbb{S}} |\mathbf{Y}| \, \mathbb{P}_{\mathcal{L}_{\mathbb{S}}}(\mathbf{Y})$$

$$= \sum_{\mathbf{Y} \subseteq \mathbb{S}} |\mathbf{Y}| \frac{\det(\mathcal{L}_{\mathbf{Y}})}{\det(\mathcal{L}_{\mathbb{S}} + \mathbf{I})}$$

where $\frac{\det(\mathcal{L}_{\mathbf{Y}})}{\det(\mathcal{L}_{\mathbb{S}} + \mathbf{I})}$ represents the probability of sampling the subset $\mathbf{Y}$, and therefore we are taking the sum over the probabilities of sampling all different $|\mathbf{Y}|$. Based on the following Lemma from Rising (2013):

> **Lemma** (Theorem 2.3.9 in Rising (2013)). *Let $\mathbf{Y} \sim \text{DPP}(\mathcal{K})$ where $\mathcal{K} = \mathbf{I} - (\mathcal{L} + \mathbf{I})^{-1}$, and let $\{\lambda_i\}_{i=1}^n$ be the eigenvalues of $\mathcal{K}$. Then $|\mathbf{Y}| = \sum_{i=1}^n Z_i$ where $\{Z_i\}_{i=1}^n$ are independent Bernoulli trials with $\mathbb{E}[Z_i] = \lambda_i$.*

we can write the expectation over the random variable $|\mathbf{Y}|$ by

$$\mathbb{E}_{\mathbf{Y} \sim \mathbb{P}_{\mathcal{L}_{\mathbb{S}}}}[|\mathbf{Y}|] = \mathbb{E}\left[\sum_{i=1}^n Z_i\right] = \sum_{i=1}^n \mathbb{E}[Z_i] = \sum_{i=1}^n \lambda_i = \text{Tr}(\mathcal{K}).$$

Since $\mathcal{K} = \mathcal{L}(\mathcal{L} + \mathbf{I})^{-1} = \mathbf{I} - (\mathcal{L} + \mathbf{I})^{-1}$ (Kulesza et al., 2012), we can relate the eigenvalues of $\mathcal{L}$ and $\mathcal{K}$ as follows,

$$\lambda_i^{\mathcal{K}} = \frac{\lambda_i^{\mathcal{L}}}{\lambda_i^{\mathcal{L}} + 1}, \quad \forall i$$

where the superscript references which matrix the eigenvalues belong to. Finally, we have,

$$\mathbb{E}_{\mathbf{Y} \sim \mathbb{P}_{\mathcal{L}_{\mathbb{S}}}}[|\mathbf{Y}|] = \sum_{i=1}^n \frac{\lambda_i^{\mathcal{L}_{\mathbb{S}}}}{\lambda_i^{\mathcal{L}_{\mathbb{S}}} + 1} = \text{Tr}\left(\mathbf{I} - (\mathcal{L}_{\mathbb{S}} + \mathbf{I})^{-1}\right)$$

and we have the expected cardinality of a sample from our DPP based upon the kernel matrix $\mathcal{L}_{\mathbb{S}}$. ∎

## D.2. Proposition 5 [Maximum Diversity]

**Proposition 5** (**Maximum Diversity**). *The diversity of a population $\mathbb{S}$ is bounded by $\text{Diversity}(\mathbb{S}) \leq \text{rank}(\mathbf{M})$, and if $\mathbf{M}$ is normalised (i.e., $\|\mathbf{M}_{[i,:]}\| = 1, \forall i$), we have $\text{Diversity}(\mathbb{S}) \leq \text{rank}(\mathbf{M})/2$. In both cases, maximal diversity is reached if and only if $\mathbf{M}$ is orthogonal.*

**Proof.** We start from the the case of an un-normalised meta-game. We can always do the SVD decomposition of $\mathbf{M}$ and get

$$\mathbf{M} = \mathbf{U}\Sigma\mathbf{V}^{\top}$$

Then, the kernel $\mathcal{L}$ of G-DPP can be written as

$$\mathcal{L} = \mathbf{M}\mathbf{M}^{\top} = (\mathbf{U}\Sigma\mathbf{V}^{\top})(\mathbf{U}\Sigma\mathbf{V}^{\top})^{\top} = \mathbf{U}\Sigma^2\mathbf{U}^{\top}$$

This means that the eigenvalues of $\mathcal{L}$ are $\lambda_i = \sigma_i^2 > 0$ where $\sigma_i$ are the entries of the diagonal of $\Sigma$. Thus, based on Proposition 4, we can write the expected cardinality given kernel $\mathcal{L}$ as

$$\mathbb{E}_{\mathbf{Y} \sim \mathbb{P}_{\mathcal{L}_{\mathbb{S}}}}[|\mathbf{Y}|] = \sum_i \frac{\lambda_i}{\lambda_i + 1} = \sum_i \frac{\sigma_i^2}{\sigma_i^2 + 1}.$$

This implies that all eigenvalues are positive and thus, a larger cardinality can only be achieved by either adding more eigenvalues or making the eigenvalues larger. Since we can only get a maximum of $n = \text{rank}(\mathbf{M})$ non-zero eigenvalues, making the eigenvalues larger will increase the terms of the summation by $\frac{\lambda_i}{\lambda_i + 1} < 1$ which means that

$$\sup \mathbb{E}_{\mathbf{Y} \sim \mathbb{P}_{\mathcal{L}_{\mathbb{S}}}}[|\mathbf{Y}|] = \text{rank}(\mathbf{M})$$

Thus, the maximum achievable diversity is obtained when population $\mathbb{S}$ makes $\mathbf{M}$ full rank and $\lambda_i \to \infty, \forall i$.

We now prove the case for a normalised (i.e., $||\mathbf{M}_{[i,:]}|| = 1, \forall i$) meta-game. We first show that a DPP is maximised over orthogonal feature vectors. To do this we show that the probability of sampling the whole set of agents is maximised for orthogonal vectors. Starting by noting Hadamard's Inequality, it states that if $\mathbf{N}$ is the matrix having columns $v_i$, then

$$\left| \det(\mathbf{N}) \right| \leq \prod_{i=1}^{n} \|v_i\|,$$

where notably equality holds if and only if the vectors $v_i$ are orthogonal. Therefore, by this inequality we also know that the determinant of $\mathbf{M}$ is maximised when the payoffs are orthogonal for each agent, as this allows the equality to hold.

Additionally, as we define our $\mathbf{M}$ to be a square matrix of size $M \times M$, we can decompose the determinant of $\mathcal{L}$ as,

$$\det(\mathcal{L}_\mathcal{Y}) = \det(\mathbf{M}\mathbf{M}^\top) = \det(\mathbf{M}) \det(\mathbf{M}^\top) = \det(\mathbf{M})^2$$

and therefore we know that $\det(\mathcal{L}_\mathcal{Y})$ is maximised whenever $\det(\mathbf{M})$ is maximised. According to the following equation,

$$\mathbb{P}_\mathcal{L}(\mathbf{Y} = \mathcal{Y}) \propto \det(\mathcal{L}_\mathcal{Y})$$

we therefore have that the probability of sampling the whole ground set is maximised when the $\det(\mathcal{L}_\mathcal{Y})$ is maximised, which coincides with the orthogonality of $\mathbf{M}$.

Finally, based on the fact that orthogonal feature vectors maximise the diversity of a population, we show that a meta-game with orthogonal feature vectors will receive an expected cardinality value of $\frac{n}{2}$.

We know that the entries of our kernel $\mathcal{L}$ are simply $L_{ij} = \boldsymbol{w}_i \boldsymbol{w}_j^\top$. As long as $||\boldsymbol{w}_i|| = 1, \forall i$ it will be the case that $L_{ii} = 1$. Additionally, as we have stated that the rows of the meta game are orthogonal, $\boldsymbol{w}_i \boldsymbol{w}_j^\top = 0$, $\forall i, j$, so all off-diagonal entries $L_{ij} = 0$. For example, the kernel $\mathcal{L}$ for a meta game of size 3 would be,

$$\mathcal{L} = \begin{bmatrix} 1 & 0 & 0 \\ 0 & 1 & 0 \\ 0 & 0 & 1 \end{bmatrix},$$

which is obviously the identity matrix of size M. The identity matrix has M eigenvalues of 1, and therefore we have

$$\mathbb{E}_{\mathbf{Y} \sim \mathbb{P}_{\mathcal{L}_\mathbb{S}}}[|\mathbf{Y}|] = \sum_{i=1}^{n} \frac{\lambda_i^\mathcal{L}}{\lambda_i^\mathcal{L} + 1} = \sum_{i=1}^{n} \frac{1}{2} = \frac{n}{2}.$$

As the expected cardinality is maximised for orthogonal feature vectors, and we know that orthogonal feature vectors will return an expected cardinality of $\frac{n}{2}$, we know that when our feature vectors are normalised this is the max achievable expected cardinality. ∎

### D.3. Proposition 6 [Diversity vs. Matrix Norm]

**Proposition 6** (**Diversity vs. Matrix Norm**). *Maximising the diversity in Eq. (11) also maximises the Frobenius norm of $\|\mathbf{M}\|_F$, but NOT vice versa.*

**Proof.** Based on Proposition 5, the maximum achievable diversity is:

$$\sup \mathbb{E}_{\mathbf{Y} \sim \mathbb{P}_{\mathcal{L}_\mathbb{S}}}[|\mathbf{Y}|] = \sup \sum_i \frac{\lambda_i}{\lambda_i + 1} = \operatorname{rank}(\mathbf{M}) = n$$

Since the terms in the above summation are $0 \leq \frac{\lambda_i}{\lambda_i + 1} < 1$, the expected cardinality is also maximised for $\lambda_i \to \infty, \forall n$. Therefore, we have that the maximum achievable diversity is attained when the eigenvalues of $\mathcal{L}$ are maximised and that $\operatorname{rank}(\mathbf{M}) = n$. In a G-DPP, we have that the kernel matrix is defined as $\mathcal{L} = \mathbf{M}\mathbf{M}^\top$, and the Frobenius norm is written as

$$\|\mathbf{M}\|_F = \sqrt{\sum_{i=1}^{n} \sum_{j=1}^{n} |\mathbf{M}_{ij}|^2} = \sqrt{\operatorname{Tr}(\mathbf{M}\mathbf{M}^\top)}$$

$$= \sqrt{\operatorname{Tr}(\mathcal{L})}$$

$$= \sqrt{\sum_{i=1}^{n} \lambda_i}$$

We have shown above that maximising the expected cardinality of $\mathcal{L}$ is achieved by maximising the eigenvalues of $\mathcal{L}$ and it

directly follows that:

$$\sup \left\| \mathbf{M} \right\|_F = \sup \sqrt{\sum_{i=1}^{n} \lambda_i}$$

is achieved as $\lambda_i \to \infty, \forall i$.

Now we allude to why maximising $||\mathbf{M}||_F$ will not necessarily lead to a large diversity of agents (i.e., the opposite direction of this proposition is not necessarily correct), but our measure based upon expected cardinality will. Firstly, note that given

$$||\mathbf{M}||_F = \sqrt{\sum_{i=1}^{n} \lambda_i},$$

an optimiser may go about increasing this function in two manners. One manner may be to continually add new eigenvalues which will help improve the overall diversity of $\mathbf{M}$. However, an optimiser may also increase this norm by focusing on one eigenvalue, or a subset of eigenvalues, and increase those to be as large as possible. Our major worry here would be that by focusing on a subset of eigenvalues, new agents that are added to the population may be redundant as they are added to increase the current set of eigenvalues - not to bring in new non-zero eigenvalues. This would lead to a lack of diversity in the population. On the other hand, note the formula for expected cardinality is,

$$\mathbb{E}_{\mathbf{Y} \sim \mathbb{P}_{\mathcal{L}_\mathbb{S}}}[|\mathbf{Y}|] = \sum_i \frac{\lambda_i}{\lambda_i + 1}$$

and note how each eigenvalue can at most contribute a value of 1 to the expected cardinality value. The importance of this is that an optimiser will gain less marginal benefit from increasing an already large eigenvalue than it would from adding a new non-zero eigenvalue to the meta game. Therefore, whilst we can guarantee that our expected cardinality measure will always search for new non-zero eigenvalues (and by that notion, more diverse agents), we can not guarantee the same for the Frobenius norm measure. ∎

### D.4. Proposition 7 [Uniqueness of Diverse Best Response]

**Proposition 7** (**Uniqueness of Diverse Best Response**). *Eq. (11) is a strictly concave function. The resulting best response in Eq. (12) has a unique solution.*

*Proof.* We study the sign of the second derivative of the diversity measure (Eq.(11)) in a neighbourhood of the positive semidefinte symmetric matrix $\mathcal{L}$. We apply a perturbation to $\mathcal{L}$ such that $\tilde{\mathcal{L}} = \mathcal{L} + \varepsilon \mathbf{A}$ with $\mathbf{A}$ being a symmetric matrix and $\varepsilon \in \mathbb{R}$, as a result, what we need to show is:

$$\frac{\partial^2}{\partial \varepsilon^2} \operatorname{Tr}\left(\mathbf{I} - (\tilde{\mathcal{L}} + \mathbf{I})^{-1}\right) \bigg|_{\varepsilon=0} < 0$$

First, we notice that

$$\mathbf{I} - (\tilde{\mathcal{L}} + \mathbf{I})^{-1} = \mathbf{I} - \sum_{n=0}^{\infty} (-1)^n (\tilde{\mathcal{L}})^n = -\sum_{n=1}^{\infty} (-1)^n (\tilde{\mathcal{L}})^n = \sum_{n=0}^{\infty} (-1)^n (\mathcal{L} + \varepsilon \mathbf{A})^{n+1}$$

where we used the matrix expansion $(\mathbf{I} + \tilde{\mathcal{L}})^{-1} = \sum_{n=0}^{\infty} (-1)^n \tilde{\mathcal{L}}^n$. We can always ensure $||\tilde{\mathcal{L}}||_F < 1$ by choosing $\varepsilon$ small enough and redefining $\mathcal{L} := \frac{\mathcal{L}}{||\mathcal{L}||_F + \beta}, \beta > 0 \in \mathbb{R}$ if necessary. We note that after this modification $\mathcal{L}$ remains a legitimate matrix for a DPP due to being positive semidefinite, and do not affect the rankings of agents. Then,

$$\begin{aligned}
\frac{\partial^2}{\partial \varepsilon^2} \left(\mathbf{I} - (\tilde{\mathcal{L}} + \mathbf{I})^{-1}\right) \bigg|_{\varepsilon=0} &= \frac{\partial^2}{\partial \varepsilon^2} \sum_{n=0}^{\infty} (-1)^n (\mathcal{L} + \varepsilon \mathbf{A})^{n+1} \bigg|_{\varepsilon=0} \\
&= \sum_{n=0}^{\infty} (-1)^n \frac{\partial^2}{\partial \varepsilon^2} (\mathcal{L} + \varepsilon \mathbf{A})^{n+1} \bigg|_{\varepsilon=0} \\
&= \frac{1}{2} \mathbf{A}^2 \sum_{n=0}^{\infty} (-1)^n (n+1)(n) \mathcal{L}^{n-1} + \frac{1}{2} \sum_{n=0}^{\infty} (-1)^n (n+1)(n) \mathcal{L}^{n-1} \mathbf{A}^2 \quad (1)
\end{aligned}$$

The series $\sum_{n=0}^{\infty}(-1)^n(n+1)(n)\mathcal{L}^{n-1}$ can be written as

$$\sum_{n=0}^{\infty}(-1)^n(n+1)(n)\mathcal{L}^{n-1} = \sum_{n=0}^{\infty}(-1)^{n+1}(n+2)(n+1)\mathcal{L}^n$$

$$= -\sum_{n=0}^{\infty}(-1)^{n+2}\frac{\partial^2}{\partial\mathcal{L}^2}\mathcal{L}^{n+2}$$

$$= -\frac{\partial^2}{\partial\mathcal{L}^2}\sum_{n=0}^{\infty}(-1)^{n+2}\mathcal{L}^{n+2}$$

$$= -\frac{\partial^2}{\partial\mathcal{L}^2}\left(\mathcal{L}^2 - \mathcal{L}^3 + \mathcal{L}^4 - \ldots\right)$$

$$= -\frac{\partial^2}{\partial\mathcal{L}^2}\left((\mathbf{I}+\mathcal{L})^{-1} + \mathcal{L} - \mathbf{I}\right)$$

$$= -2(\mathbf{I}+\mathcal{L})^{-3}$$

where we used the matrix expansion $(\mathbf{I}+\mathcal{L})^{-1} = \sum_{n=0}^{\infty}(-1)^n\mathcal{L}^n$ which holds under the modifications specified before.

Finally, we obtain that

$$\frac{\partial^2}{\partial\varepsilon^2}\operatorname{Tr}\left(\mathbf{I} - (\tilde{\mathcal{L}}+\mathbf{I})^{-1}\right)\bigg|_{\varepsilon=0} = \operatorname{Tr}\left(\frac{\partial^2}{\partial\varepsilon^2}\left(\mathbf{I} - (\tilde{\mathcal{L}}+\mathbf{I})^{-1}\right)\right)\bigg|_{\varepsilon=0}$$

$$= -\operatorname{Tr}\left(\mathbf{A}^2(\mathbf{I}+\mathcal{L})^{-3} + (\mathbf{I}+\mathcal{L})^{-3}\mathbf{A}^2\right)$$

$$< 0$$

Where the last inequality comes from the fact that both $(\mathbf{I}+\mathcal{L})^{-3}$ and $\mathbf{A}^2$ are positive definite and the trace of the product of two positive definite matrices is positive. We therefore have shown that the expected cardinality is a strictly concave function, which are known to have a unique optimiser. ∎

### D.5. Theorem 8 [Convergence of Diverse FP]

**Theorem 8** (**Convergence of Diverse FP**). *The perturbation sequence induced by diverse FP process is a uniformly bounded martingale difference sequence; therefore, diverse FP shares the same convergence property as GWFP.*

**Proof.** *We begin by proving that our perturbation sequence is a martingale sequence w.r.t. the normalised expected cardinality (this has no effect on rankings of agents),*

$$X_n = \frac{\mathbb{E}_{\mathbf{Y}\sim\mathbb{P}_{\mathcal{L}_\mathbb{S}}}[|\mathbf{Y}|]}{|\mathcal{Y}|}$$

*Based on Proposition 5, assuming a normalised payoff table, we know $0 \leq X_n \leq 0.5$. We are to show that $S_n = \frac{1}{n}\sum_{i=1}^{n}[X_i - \mu]$ where $\mu = \mathbb{E}[X_n]$ is a martingale sequence by proving that it meets the three conditions listed below one by one.*

1. *$S_n$ is a measurable function of $X_1, X_2, ..., X_n$*

   *$S_n$ is a measurable function as it is a partial sum of $\{X_i\}_{i=1}^{\infty}$.*

2. *$\mathbb{E}[|S_n|] < \infty, \forall n$*

$$\mathbb{E}[|S_n|] \leq \frac{1}{n}\sum_{i=1}^{n}\mathbb{E}[|X_i - \mu|] < \infty$$

3. $\mathbb{E}[S_{n+1}|X_1,...X_n] = S_n$

$$\begin{aligned}\mathbb{E}[S_{n+1}|X_1,...,X_n] &= \mathbb{E}[S_n + \frac{1}{n}X_{n+1} - \frac{1}{n}\mu] \\ &= S_n + \mathbb{E}[\frac{1}{n}X_{n+1} - \frac{1}{n}\mu|X_1,...,X_n] \\ &= S_n - \frac{1}{n}\mu + \mathbb{E}[\frac{1}{n}X_{n+1}|X_1,...,X_n] \\ &= S_n - \frac{1}{n}\mu + \frac{1}{n}\mu \\ &= S_n\end{aligned}$$

and therefore we have that $\{S_k\}_{k=1}^{\infty}$ is a martingale sequence w.r.t. $\{X_k\}_{k=1}^{\infty}$. Now we let $M_k = S_k - S_{k-1}, k = 2,3,...$ and show that this is a martingale **difference** sequence under the same three conditions.

1. $M_k$ is a measurable function of $X_1, X_2, ..., X_n$

$M_k$ is measurable as both $S_k$ and $S_{k-1}$ are.

2. $\mathbb{E}[|M_k|] < \infty, \quad \forall k$

$$\mathbb{E}[|M_k|] \le \mathbb{E}[|S_k|] + \mathbb{E}[|S_{k-1}|] < \infty$$

3. $\mathbb{E}[M_{k+1}|X_1,...,X_n] = 0$

$$\begin{aligned}\mathbb{E}[M_{k+1}|X_1,...,X_n] &= \mathbb{E}[S_{k+1} - S_k|X_1,...,X_n] \\ &= \mathbb{E}[S_{k+1}|X_1,...,X_n] - S_k \\ &= S_k - S_k \\ &= 0\end{aligned}$$

and we have shown that $\{M_k\}_{k=1}^{\infty}$ is a martingale difference sequence w.r.t. $\{X_k\}_{k=1}^{\infty}$.
Next, we show that this martingale sequence is bounded uniformly in L2. $M_n$ is said to be bounded in L2 if:
$$\sup_n[M_n^2] < \infty$$
which can be shown due to the following:
$$\sup_n\left[M_n^2\right] = \sup_n\left[\frac{1}{n}\sum_{i=1}^{n}\left[X_i - \mu\right]^2\right] < \infty$$
$$as \; \frac{1}{n}\sum_{i=1}^{n}\left[X_i - \mu\right]^2 \le 0.25, \forall n$$

where the last line holds due to $0 \le X_i \le 0.5$. Therefore, if $M_n \in L2$, then the martingale sequence $M$ is bounded in L2 if and only if:
$$\sum_{k\ge 1}\mathbb{E}\left[(M_k - M_{k-1})^2\right] < +\infty$$
which we can show in the following manner:
$$\begin{aligned}\sum_{k\ge 1}\mathbb{E}\left[(M_k - M_{k-1})^2\right] &= \sum_{k\ge 1}\left[\mathbb{E}[M_k^2 - 2M_kM_{k-1} + M_{k-1}^2]\right] \\ &= \sum_{k\ge 1}\left[\mathbb{E}[M_k^2] + \mathbb{E}[M_{k-1}^2] - \mathbb{E}[2M_kM_{k-1}]\right] \\ &= \sum_{k\ge 1}\left[\mathbb{E}[M_k^2] + \mathbb{E}[M_{k-1}^2]\right] \\ &< \infty\end{aligned}$$

as $M_k \in L2, \forall k$. *where we note that* $\mathbb{E}[2M_k M_{k-1}] = 0$ *due to the martingale difference sequence properties of* $M_k$. *Finally, as we satisfy both of the conditions of GWFP, namely that our expected cardinality function is strictly concave as shown in Proposition 7 and the perturbation meets the condition of Eq. (5) we have that as* $\tau \to 0$ *our diverse FP process will almost surely result in a GWFP process, which are known to converge in two-player zero-sum games and potential games (Leslie & Collins, 2006).* ∎

### D.6. Proposition 9 [Gamescape Enlargement]

**Proposition 9** (**Gamescape Enlargement**). *Adding a new best-response policy* $S_{\boldsymbol{\theta}}$ *via Eq. (13) strictly enlarges the gamescape. Formally, we have*

$$\text{Gamescape}\left(\mathbb{S}\right) \subsetneq \text{Gamescape}\left(\mathbb{S} \cup \{S_{\boldsymbol{\theta}}\}\right)$$

**Proof.** *According to Proposition 5 and Proposition 7, maximising the expected cardinality of the DPP implies there exists a unique solution that makes the meta-game payoff table* **M** *full rank and increases the eigenvalues. If* **M** *is full rank, this means that the new row in the meta game* **M** *corresponding to the new policy* $S_{\boldsymbol{\theta}}$ *must be linearly independent to the other rows and thus it must not be a convex combination of the other rows, as a result, the gamescape is strictly enlarged. Such a result is also expected because the diversity in terms of expected cardinality promotes orthogonality, which is a stronger property than linear independency.* ∎

### D.7. Theorem 10 [Convergence of Diverse $\alpha$-PSRO]

**Theorem 10** (**Convergence of Diverse $\alpha$-PSRO**). *Diverse $\alpha$-PSRO with the Oracle of Eq. (14) converges to the sub-cycle of the unique SSCC in the two-player symmetric games.*

**Proof.** *Our proof follows the same argument as (Muller et al., 2019) Proposition 3, which can be adapted in the content of our diverse PBR oracle in Eq. (14). The uniqueness of the SSCC follows from the fact that in the single-population case, the response graph is fully-connected. Suppose at termination of $\alpha$-DPP, the $\alpha$-population contains no strategy within the SSCC, and let $s$ be a strategy in the $\alpha$-DPP population. We claim that $s$ attains a higher value for the objective defining the $\alpha$-DPP oracle than any strategy in the $\alpha$-DPP population, which contradicts the fact that $\alpha$-DPP has terminated. By virtue of being in the SSCC we have that $M^1(s, s') > M^2(s', s)$ for all $s'$ outside the SSCC, and in particular for all $s' \in \mathbb{S}_t^i$, thus the PBR objective for $s$ is 1.*

*If a member of the underlying game's SSCC appears in the $\alpha$-DPP population, this member will induce its own meta-SSCC in the meta game's response graph which will have positive probability under the $\alpha$-Rank distribution for the meta-game, and the DPP oracle for this meta-SSCC will always return a member of the underlying game's SSCC. This is because the only strategies that receive a non-zero quality score are those that are members of the underlying game's SSCC, and we ignore strategies with a quality of zero. If the only strategy that returns a non-zero quality score, which must be a member of the SSCC, is already in the population, the corresponding meta-SSCC already contains a cycle of the underlying SSCC. Note that if the meta-SSCC does not contain a cycle, it must be a singleton. Either this singleton is equal to the full SSCC of the underlying game (in which we have $\alpha$-fully converged), or it is not, in which case the DPP oracle must return a new strategy from the underlying SSCC, contradicting our assumption that is has terminated.* ∎

# E. Additional Results on Real-World Meta-Games

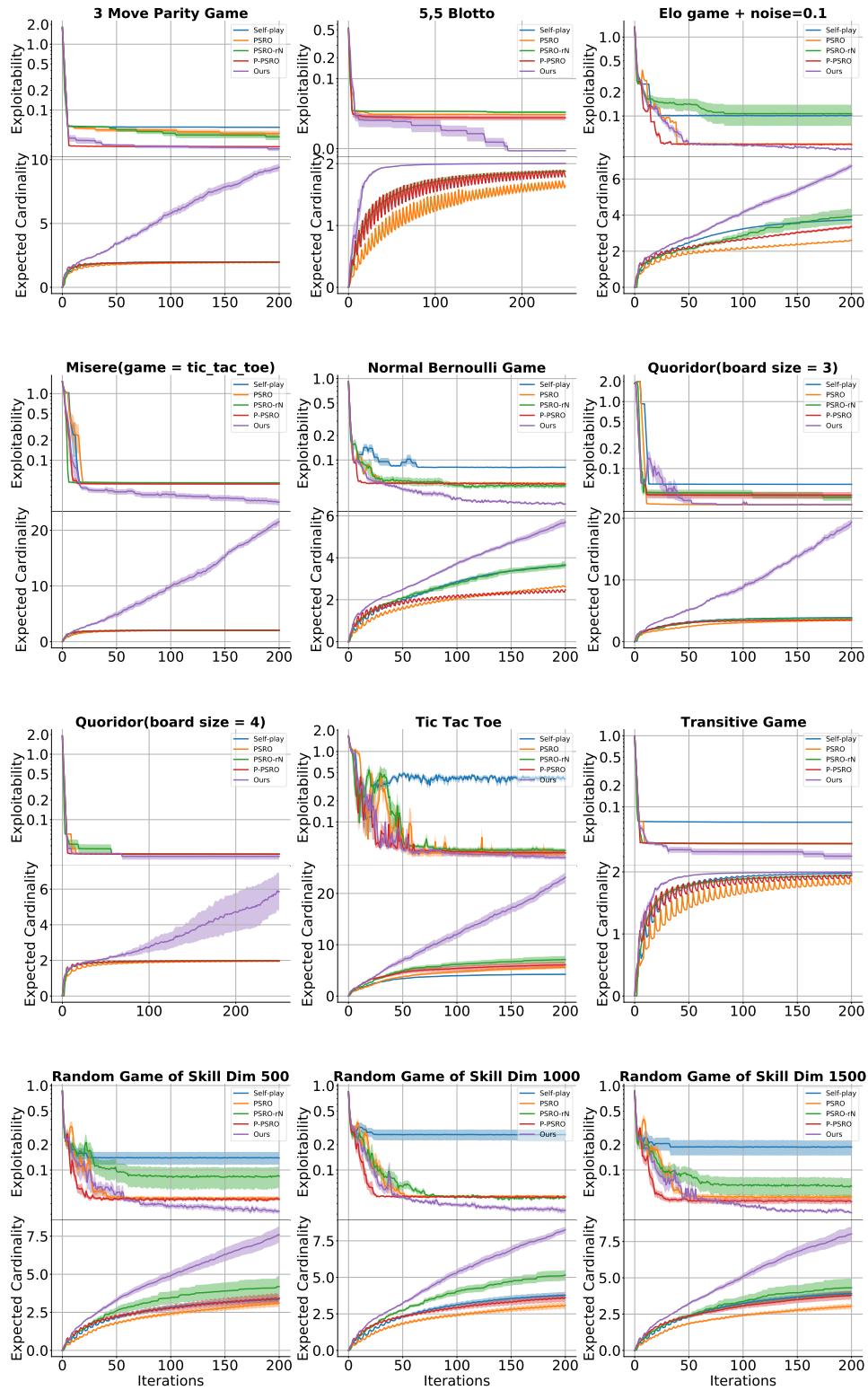

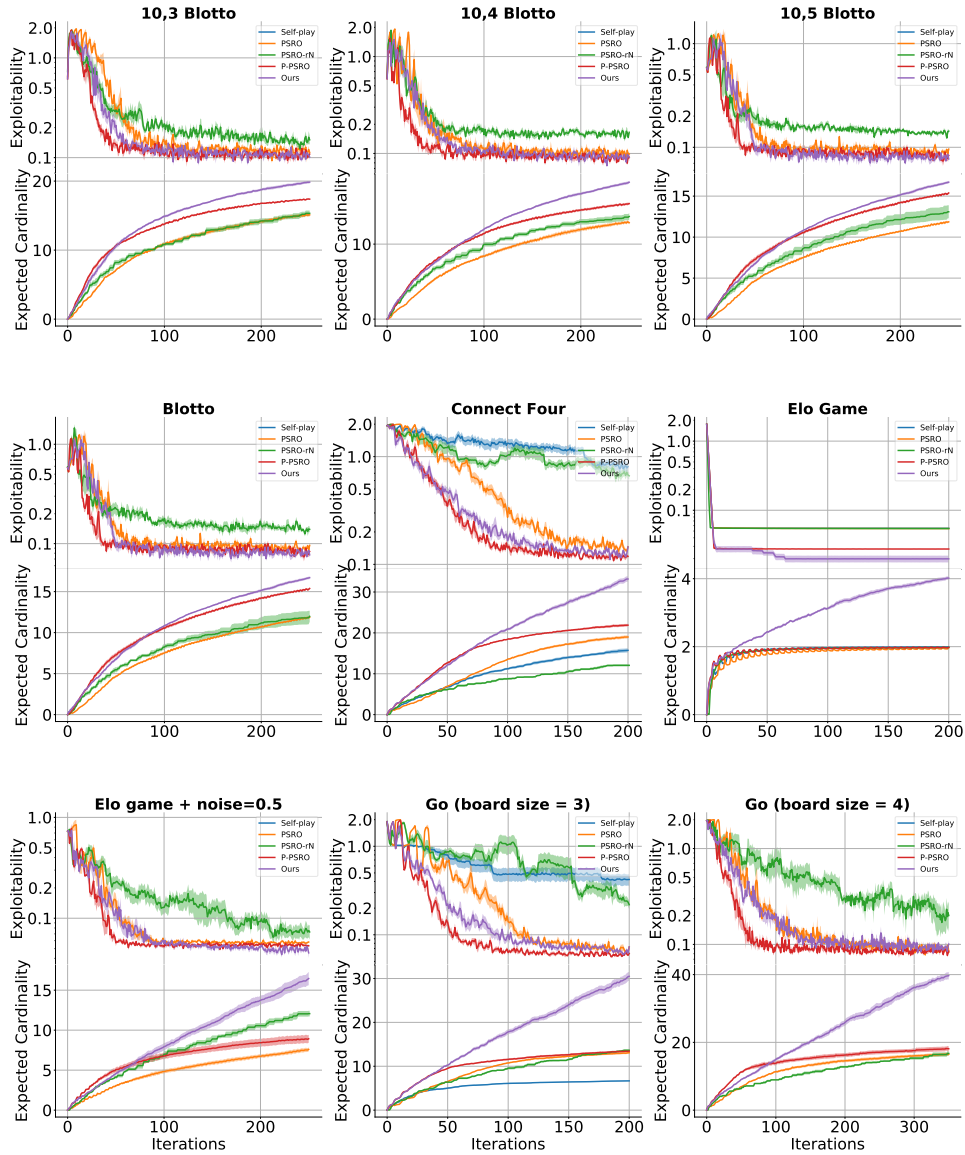

## F. Empirical Result of Frobenius Norm vs. Expected Cardinality

We illustrate here a counter-example of when replacing our expected cardinality term in Eq. (13) with a Frobenius norm measure on the meta-game **M** over the population $\mathbb{S}^1 \cup \{S_\theta\}$ will fail to improve both the expected cardinality of **M** and so be unable to break the exploitability plateau of PSRO.

Specifically, we use a 150 dimensional Random Game of Skill (Czarnecki et al., 2020) where 25 of the strategies in the underlying game are particularly bad. In this setting, the Frobenius norm struggles as the poor-performing strategies all induce a large norm, and are therefore selected to be added to $\mathbb{S}^1$. On the other hand, at worst our expected cardinality measure will select one of these poor-performing strategies, but the remaining poor strategies are no longer diverse and are not selected. Therefore, whilst we suspect the Frobenius norm measure can work as a diversity measure, it is not as general as our metric and can fail under some circumstances, i.e. poor-performing strategies in the underlying game.

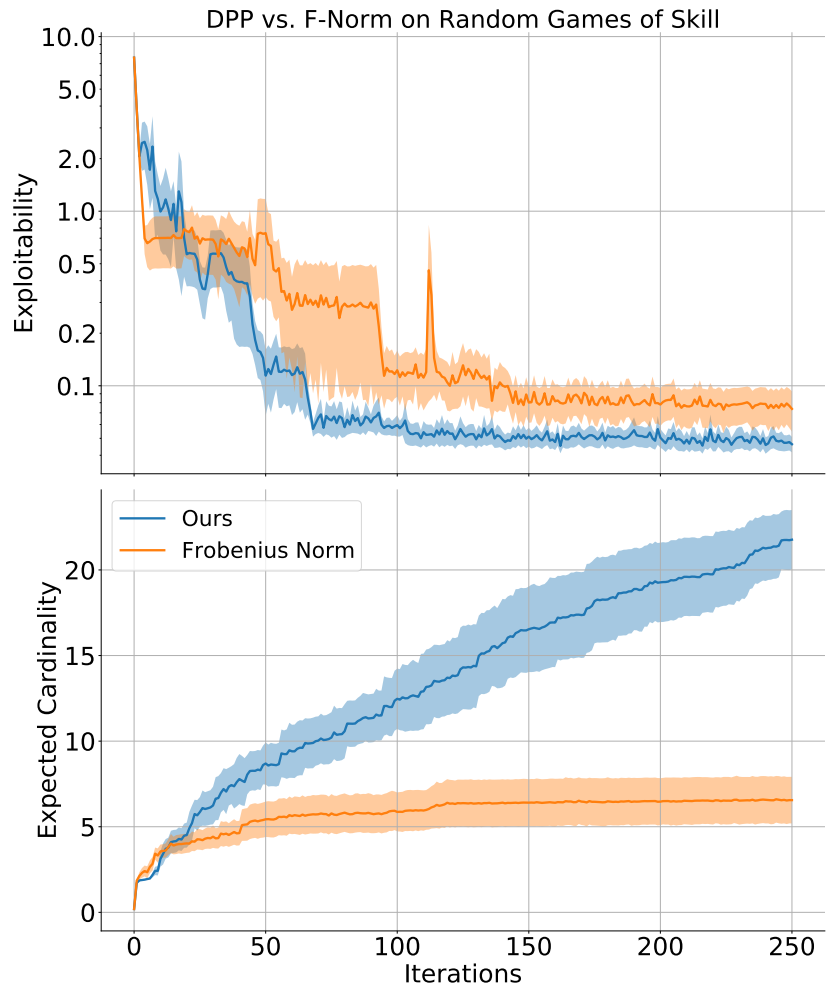

# G. Hyper-parameter Settings for All Experiments

The code can be found in: https://github.com/oslumbers/diverse-psro

Table 1. Hyper-parameter settings for our diverse-PSRO method *vs.* other baselines on four experiments.

| SETTINGS | VALUE | DESCRIPTION |
|---|---|---|
| **REAL-WORLD META-GAMES** | | |
| ORACLE METHOD | DIVERSE BEST RESPONSE | SUBROUTINE OF GETTING ORACLES |
| LEARNING RATE | 0.5 | LEARNING RATE FOR AGENTS |
| IMPROVEMENT THRESHOLD | 0.03 | CONVERGENCE CRITERIA |
| METASOLVER | FICTITIOUS PLAY | METASOLVER METHOD |
| METASOLVER ITERATIONS | 1000 | # ITERATIONS FOR METASOLVER |
| # OF THREADS IN PIPELINE | 2 | # LEARNERS IN PIPELINE PSRO |
| # OF SEEDS | 10 | # TRIALS |
| DPP WEIGHTING | 0.15 | WEIGHT OF THE DPP BEST RESPONSE |
| **NON-TRANSITIVE MIXTURE MODEL** | | |
| ORACLE | GRADIENT ASCENT | SUBROUTINE OF GETTING ORACLES |
| OPTIMIZER | ADAM | GRADIENT ASCENT OPTIMIZER |
| LEARNING RATE | 0.1 | LEARNING RATE FOR OPTIMIZER |
| BETAS | (0.9, 0.99) | BETAS PARAMETER FOR OPTIMIZER |
| $\pi^i$ | $\pi_k^i = \exp\left(-(x_i - \mu_k)^\top \Sigma (x_i - \mu_k)/2\right)$ | STRATEGY FROM 2D COORDINATES |
| GAME PAYOFF | $\pi^{1,\top} \begin{bmatrix} 0 & 1 & 1 & 1 & -1 & -1 & -1 \\ -1 & 0 & 1 & 1 & 1 & -1 & -1 \\ -1 & -1 & 0 & 1 & 1 & 1 & -1 \\ -1 & -1 & -1 & 0 & 1 & 1 & 1 \\ 1 & -1 & -1 & -1 & 0 & 1 & 1 \\ 1 & 1 & -1 & -1 & -1 & 0 & 1 \\ 1 & 1 & 1 & -1 & -1 & -1 & 0 \end{bmatrix} \pi^2 + \frac{1}{2}\sum_{k=1}^{7}(\pi_k^1 - \pi_k^2)$ | PAYOFF FOR $\pi^1$ AGAINST $\pi^2$ |
| $\Sigma$ | $1/2 I$ | COVARIANCE MATRIX FOR GAUSSIANS |
| $\mu_1$ | $(2.872, -0.025)$ | POSITION OF THE FIRST GAUSSIAN |
| $\mu_2$ | $(1.8105, 2.2298)$ | POSITION OF THE SECOND GAUSSIAN |
| $\mu_3$ | $(1.8105, -2.2298)$ | POSITION OF THE THIRD GAUSSIAN |
| $\mu_4$ | $(-0.61450, 2.8058)$ | POSITION OF THE FOURTH GAUSSIAN |
| $\mu_5$ | $(-0.61450, -2.8058)$ | POSITION OF THE FIFTH GAUSSIAN |
| $\mu_6$ | $(-2.5768, 1.2690)$ | POSITION OF THE SIXTH GAUSSIAN |
| $\mu_7$ | $(-2.5768, -1.2690)$ | POSITION OF THE SEVENTH GAUSSIAN |
| STRATEGY INITIALISATION VARIANCE | $0.1^2$ | VARIANCE OF GAUSSIAN DISTRIBUTION AT 0 |
| METASOLVER | FICTITIOUS PLAY | METASOLVER METHOD |
| METASOLVER ITERATIONS | 1000 | # ITERATIONS FOR METASOLVER |
| ITERATIONS | 50 | # TRAINING ITERATIONS |
| DPP WEIGHT AT ITERATION $t$ | $\dfrac{0.7}{1 + \exp(-0.25(t - 25))}$ | WEIGHT OF THE DPP BEST RESPONSE |
| # OF THREADS IN PIPELINE | 4 | # LEARNERS IN PIPELINE PSRO |
| # OF SEEDS | 10 | # OF TRIALS |
| **COLONEL BLOTTO** | | |
| ORACLE METHOD | ZERO-ORDER | SUBROUTINE OF GETTING ORACLES |
| ORACLE ITERATIONS | 50 | NEW STRATEGIES PER ITERATION |
| $\mu$ | 0.1 | LEARNING RATE |
| METASOLVER | LINEAR PROGRAMMING | METASOLVER METHOD |
| # OF SEEDS | 10 | # OF TRIALS |
| DPP WEIGHT | 0.15 | WEIGHT OF THE DPP BEST RESPONSE |
| # OF AREAS | 3 | # OF AREAS TO DISTRIBUTE COINS |
| # OF COINS | 10 | # OF COINS TO DISTRIBUTE |
| **$\alpha$-PSRO** | | |
| ORACLE METHOD | PBR / DPP-PBR | SUBROUTINE OF GETTING ORACLES |
| METASOLVER | $\alpha$-RANK | METASOLVER METHOD |
| ITERATIONS | 50 | # OF TRAINING ITERATIONS |
| # OF SEEDS | 20 | # OF TRIALS |
| $\alpha$ | INFINITE | THE $\alpha$ IN $\alpha$-RANK |

# H. Implementation of Oracles

## H.1. Pseudocodes

---
**Algorithm 2** Diverse Best Response Oracle

---
1: **Inputs:**
2: Player Populations $\mathbb{S}_t = \prod_{i \in \mathcal{N}} \mathbb{S}_t^i$, with $S_t^i \in \mathbb{S}_t^i$ parametrised by $\theta_{S_t^i}$
3: Metapolicies $\boldsymbol{\pi}_t = \prod_{i \in \mathcal{N}} \boldsymbol{\pi}_t^i$
4: Learning rate $\mu$
5: Diversity probability $\lambda$
6:
7: **function** oracle($\mathbb{S}_t, \boldsymbol{\pi}_t, \mu, \lambda$)
8:     Compute $\mathbf{BR}_{qual} = \mathbf{BR}^i(\mathbb{S}_t^{-i}, \boldsymbol{\pi}_t^{-i})$
9:     **for** each pure strategy $P_j$ **do**:
10:         Update meta-payoff $\mathbf{M}_j = \mathbf{M}(\mathbb{S}_t^i \cup \{P_j\})$
11:     Compute $\mathbf{BR}_{div} = \arg\max_{P_j} \left( \mathrm{Tr} \left( \mathbf{I} - (\mathbf{M}_j \mathbf{M}_j^\top + \mathbf{I})^{-1} \right) \right)$.
12:     Choose $\mathbf{BR} = \mathbf{BR}_{div}$ with probability $\lambda$ else $\mathbf{BR} = \mathbf{BR}_{qual}$
13:     Update $\theta_{S_t^i} = \mu \theta_{S_t^i} + (1-\mu)\theta_{\mathbf{BR}}$
14:     **Return:** $S_t^i$

---

---
**Algorithm 3** Diverse Gradient Ascent Oracle

---
1: **Inputs:**
2: Player Populations $\mathbb{S}_t = \prod_{i \in \mathcal{N}} \mathbb{S}_t^i$, with $S_t^i \in \mathbb{S}_t^i$ parametrised by $\theta_{S_t^i}$
3: Metapolicies $\boldsymbol{\pi}_t = \prod_{i \in \mathcal{N}} \boldsymbol{\pi}_t^i$
4: Number of training updates $N_{train}$
5: Diversity weight $\lambda$
6:
7: **function** oracle($\mathbb{S}_t, \boldsymbol{\pi}_t, N_{train}, \lambda$)
8:     Randomly initialise a new $S^{train}$
9:     **for** $j = 1, \ldots, N_{train}$:
10:         Compute payoff $p_j$ of $S^{train}$
11:         Compute meta-payoff $\mathbf{M}_j = \mathbf{M}(\mathbb{S}_t^i \cup \{S^{train}\})$
12:         Compute diversity $d_j = \mathrm{Tr}\left(\mathbf{I} - (\mathbf{M}_j \mathbf{M}_j^\top + \mathbf{I})^{-1}\right)$
13:         Compute loss $l_j = -(1-\lambda)p_j - \lambda d_j$
14:         Update $\theta_{S^{train}}$ to minimise $l_j$ using a gradient based optimization method
15:     **Return:** $S^{train}$

---

**Algorithm 4** Diverse Zero-order Oracle

1: **Inputs:**
2: Player Populations $\mathbb{S}_t = \prod_{i \in \mathcal{N}} \mathbb{S}_t^i$, with $S_t^i \in \mathbb{S}_t^i$ parametrised by $\theta_{S_t^i}$
3: Metapolicies $\boldsymbol{\pi}_t = \prod_{i \in \mathcal{N}} \boldsymbol{\pi}_t^i$
4: Learning rate $\mu$
5: Noise parameter $\sigma$
6: Diversity weight $\lambda$
7:
8: **function** oracle($\mathbb{S}_t, \boldsymbol{\pi}_t, \mu, \sigma, \lambda$)
9:     Sample a single $S_t^i \in \mathbb{S}_t^i$ with probability $\boldsymbol{\pi}_t^i$
10:     **for** perturbation $j \in 1, 2, \ldots$ **do**:
11:         $S_j = \text{RandomPerturbation}(S_t^i, \mu, \sigma)$
12:         Update meta-payoff $\mathbf{M}_j = \mathbf{M}(\mathbb{S}_t^i \cup \{S_j\})$
13:         Compute payoff $p_j$ of $S_j$
14:         Compute diversity $d_j = \text{Tr}\left(\mathbf{I} - (\mathbf{M}_j \mathbf{M}_j^\top + \mathbf{I})^{-1}\right)$.
15:     **Return:** $\arg\max_{S_j} (1-\lambda)p_j + \lambda d_j$
16: **function** RandomPerturbation($S_t^i, \mu, \sigma$)
17:     Generate noise $\varepsilon \sim \mathcal{N}(0, \sigma^2)$
18:     Add noise $\theta_{S_t^i} = \mu \theta_{S_t^i} + (1-\mu)\varepsilon$
19:     **Return** $S_t^i$

**Algorithm 5** Diverse Reinforcement Learning Oracle

1: **Inputs:**
2: Player Populations $\mathbb{S}_t = \prod_{i \in \mathcal{N}} \mathbb{S}_t^i$, with $S_t^i \in \mathbb{S}_t^i$ parametrised by $\theta_{S_t^i}$
3: Metapolicies $\boldsymbol{\pi}_t = \prod_{i \in \mathcal{N}} \boldsymbol{\pi}_t^i$,
4: Learning rate $\mu$,
5: Noise parameter $\sigma$,
6: Diversity weight $\lambda$,
7: **function** oracle($\mathbb{S}_t, \boldsymbol{\pi}_t, \mu, \sigma, \lambda$)
8:     Use an RL algorithm to find $\mathbf{BR}_{RL} = \mathbf{BR}_{RL}(\mathbb{S}_t^{-i}, \boldsymbol{\pi}_t^{-i})$
9:     **for** perturbation $j \in 1, 2, \ldots$ **do**:
10:         $\mathbf{BR}_j = \text{RandomPerturbation}(\mathbf{BR}_{RL}, \mu, \sigma)$
11:         Update meta-payoff $\mathbf{M}_j = \mathbf{M}(\mathbb{S}_t^i \cup \{\mathbf{BR}_j\})$
12:         Compute payoff $p_j$ of $S_j$
13:         Compute diversity $d_j = \text{Tr}\left(\mathbf{I} - (\mathbf{M}_j \mathbf{M}_j^\top + \mathbf{I})^{-1}\right)$.
14:     **Return:** $\arg\max_{S_j} (1-\lambda)p_j + \lambda d_j$
15: **function** RandomPerturbation($\mathbf{BR}_{RL}, \mu, \sigma$)
16:     Generate noise $\varepsilon \sim \mathcal{N}(0, \sigma^2)$
17:     Add noise $\theta_{\mathbf{BR}} = \mu \theta_{\mathbf{BR}_{RL}} + (1-\mu)\varepsilon$
18:     **Return BR**

### H.2. Time Complexity: PSRO vs. Diverse PSRO on Normal-Form Games

**Time Complexity of PSRO on Solving Normal-Form Games**

**Input:** Candidate population $\Pi \in \mathbb{R}^{P \times D}, P \ll D$, game $\boldsymbol{G} \in \mathbb{R}^{D \times D}$

1. Compute meta-game $\boldsymbol{M} = \Pi \boldsymbol{G} \Pi^\top \rightarrow \mathcal{O}(PDP + DDP) = \mathcal{O}(P^2 D + D^2 P)$
2. Compute metanash $\pi$ of meta-game $\boldsymbol{M} \in \mathbb{R}^{P \times P} \rightarrow \mathcal{O}(P^3)$
3. Aggregate the population of strategies $q = \pi^\top \Pi \rightarrow \mathcal{O}(PD)$
4. **For** each pure strategy $p_i$ of a total of D **do**
    - Compute payoff $p_i \boldsymbol{G} q \rightarrow \mathcal{O}(D^2)$

**Then PSRO takes** $\mathcal{O}(P^2 D + D^2 P + P^3 + PD + D^3) = \mathcal{O}(D^3)$

**Time Complexity of Diverse PSRO on Solving Normal-Form Games**

**Input:** Candidate population $\Pi \in \mathbb{R}^{P \times D}, P \ll D$, game $\boldsymbol{G} \in \mathbb{R}^{D \times D}$

1. Compute meta-game $\boldsymbol{M} = \Pi \boldsymbol{G} \Pi^\top \rightarrow \mathcal{O}(PDP + DDP) = \mathcal{O}(P^2 D + D^2 P)$
2. Compute metanash $\pi$ of meta-game $\boldsymbol{M} \in \mathbb{R}^{P \times P} \rightarrow \mathcal{O}(P^3)$
3. Aggregate the population of strategies $q = \pi^\top \Pi \rightarrow \mathcal{O}(PD)$
4. **For** each pure strategy $p_i$ of a total of D **do**
    - Compute payoff $p_i \boldsymbol{G} q \rightarrow \mathcal{O}(D^2)$
    - Update training strategy resulting in $\mathbf{BR}_{div} \rightarrow \mathcal{O}(D)$
    - Compute meta-game $\boldsymbol{M} = \Pi \boldsymbol{G} \Pi^\top \rightarrow \mathcal{O}(PDP + DDP) = \mathcal{O}(P^2 D + D^2 P)$
    - Compute metanash $\pi$ of meta-game $\boldsymbol{M} \in \mathbb{R}^{P \times P} \rightarrow \mathcal{O}(P^3)$
    - Compute $\mathcal{L} = \mathbf{M}\mathbf{M}^\top \rightarrow \mathcal{O}(P^3)$
    - Compute expected cardinality of $\mathcal{L}$ in Proposition 4, the complexity is $\mathcal{O}(P^3)$

**Then Diverse PSRO takes** $\mathcal{O}(P^2 D + D^2 P + P^3 + PD + D^3 + D^2 + PD^3 + P^2 D^2 + 3P^3) = \mathcal{O}(PD^3)$


\end{document}

%% file: math_commands.tex
\usepackage{amsmath,amsfonts,bm}

\def\eqref#1{equation~\ref{#1}}

\def\1{\bm{1}}

\def\vtheta{{\bm{\theta}}}

\def\vm{{\bm{m}}}

\def\vpi{{\boldsymbol{\pi}}}
\def\vtheta{{\boldsymbol{\theta}}}

\def\mG{{\bm{G}}}

\DeclareMathAlphabet{\mathsfit}{\encodingdefault}{\sfdefault}{m}{sl}
\SetMathAlphabet{\mathsfit}{bold}{\encodingdefault}{\sfdefault}{bx}{n}
\newcommand{\tens}[1]{\bm{\mathsfit{#1}}}

\def\tM{{\tens{M}}}

\DeclareMathOperator*{\argmax}{arg\,max}

\DeclareMathOperator{\Tr}{Tr}